\documentclass{article} % For LaTeX2e
\usepackage[table]{xcolor}

\usepackage{iclr2024_conference,times}

\usepackage{amsmath,amsfonts,bm}

\def\eqref#1{equation~\ref{#1}}
\def\1{\bm{1}}

\DeclareMathAlphabet{\mathsfit}{\encodingdefault}{\sfdefault}{m}{sl}
\SetMathAlphabet{\mathsfit}{bold}{\encodingdefault}{\sfdefault}{bx}{n}

\usepackage{hyperref}
\usepackage{url}
\usepackage{times}
\usepackage{latexsym}
\usepackage{booktabs}       % professional-quality tables
\usepackage{amsfonts}       % blackboard math symbols
\usepackage{nicefrac}
\usepackage{latexsym}
\usepackage{multirow}
\usepackage{booktabs}
\usepackage{amsmath}
\usepackage{booktabs}
\usepackage{graphicx}
\usepackage{subfigure}

\usepackage{caption}
\usepackage{enumerate}
\usepackage{framed}
\usepackage{pifont}
\newcommand{\cmark}{\ding{51}}
\newcommand{\xmark}{\ding{55}}
\usepackage{relsize}
\usepackage{enumitem}
\usepackage{tcolorbox}
\usepackage{wrapfig}

\tcbuselibrary{skins, breakable, theorems}
\definecolor{mypink}{rgb}{.99,.91,.95}

\newtcbtheorem{example}{\scriptsize{Example}}%
  {enhanced, breakable,
    colback = white, colframe = gray, colbacktitle = gray,
    attach boxed title to top left = {yshift = -2mm, xshift = 5mm},
    boxed title style = {sharp corners},
    fonttitle = \sffamily\bfseries}{qst}

\title{L-Eval: Instituting Standardized Evaluation for Long Context Language Models}

\author{Chenxin An, Shansan Gong, Ming Zhong, Xingjian Zhao, Mukai Li, Jun Zhang,\\
\textbf{Lingpeng Kong, Xipeng Qiu}\\
Fudan University, The University of Hong Kong \\
University of Illinois Urbana-Champaign \\
Shanghai AI Lab
}

\iclrfinalcopy % Uncomment for camera-ready version, but NOT for submission.
\begin{document}

\maketitle

\begin{abstract}
Recently, there has been growing interest in extending the context length of large language models (LLMs), aiming to effectively process long inputs of one turn or conversations with more extensive histories. While proprietary models such as GPT-4 and Claude can largely preserve the reasoning ability in an extended context, open-source models are still progressing through the early stages of development. 
To bridge this gap, we propose L-Eval to institute a more standardized evaluation for long context language models (LCLMs) addressing two key aspects: dataset construction and evaluation metrics. On the one hand, we build a new evaluation suite containing 20 sub-tasks, 508 long documents, and over 2,000 human-labeled query-response pairs encompassing diverse question styles, domains, and input length (3k$\sim$200k tokens). On the other hand, we investigate the effectiveness in evalution metrics for LCLMs. Results show that popular n-gram matching metrics generally can not correlate well with human judgment, and thus we strongly advocate for length-instruction-enhanced (LIE) evaluation and employing LLM judges.  We conducted a comprehensive study of 4 popular commercial LLMs and 12 open-source counterparts using the L-Eval benchmark. Our empirical findings offer useful insights into the study of LCLMs and lay the groundwork for the development of more principled evaluation of these models.\footnote{We release our new evaluation suite, code, and all generation results on  \url{https://github.com/OpenLMLab/LEval}}

% show that fine-tuning longer with scaled positional embedding does offer benefits for closed-ended tasks but falls short in open-ended tasks.

% Vicuna1.5-16k trained with scaled position embedding on Llama2-4k obtain superior performance on closed-ended tasks but it fails to outperform the 4k counterpart in open-ended tasks due to a weakened understanding of instructions\footnote{We will release our suite, code, and all generation results.}. \sansa{It is better to summarize a more high-level conclusion in abstract}

\end{abstract}

\section{Introduction}
Currently, a significant amount of effort is being dedicated to research on extending the context length of large language models. Popular solutions mainly involve further pretraining or finetuning standard models on longer inputs using more efficient architectures~\citep{ding2023longnet, NEURIPS2022_67d57c32, liang2023unleashing, mohtashami2023landmark, li2023incontext}, as well as scaled positional embedding~\citep{su2022roformer, sun2022lengthextrapolatable, fixedNTK, qin2023linear}. 

There are extensive multi-task benchmarks~\citep{hendrycks2021measuring, suzgun2022challenging} for language models with short prompts, yet a high-quality one in long context modeling has not yet been established, presenting an opportunity for further development in this area.  Meanwhile, almost all previous long-sequence text generation benchmarks relied primarily on n-gram matching metrics~\citep{zhang2023cab, shaham2022scrolls}, such as ROUGE~\citep{lin-2004-rouge}. Whether these commonly used metrics correlate well with human judgment when testing LCLMs in a zero-shot setting remains a question. Furthermore, the open-source community has released a considerable number of language models with 16k, or 32k context length~\citep{longchat2023, du2022glm}. A comprehensive comparative study of these models can be of great value.

To address these issues, we propose \textit{L-Eval} to call for a more standardized evaluation of long context language models. For dataset construction, L-Eval has 20 sub-tasks, 4 sub-tasks are annotated from scratch (\S\ref{sec:scratch}), 4 sub-tasks are re-annotated from the public datasets (\S\ref{sec:re-anno}), and the remaining 12 sub-tasks are manually cleaned from previous long sequence datasets.  
We divide these tasks in L-Eval into two groups: closed-ended tasks and open-ended tasks. The closed-ended group primarily tests the reasoning and understanding ability regarding a longer context, and the open-ended group consists of more summarization tasks that require aggregation of long document information. In the design of L-Eval, we prioritize diversity and quality over quantity, ensuring correctness by manually validating all samples after data collection (\S\ref{sec:postprocess}). Our data diversity, indicative in question styles, domain selection, and input lengths, is detailed in Table~\ref{tab:datasets}.
% we prioritize diversity and quality over quantity. We ensure the quality and correctness by manually checking all these samples after data collection(\S\ref{sec:postprocess}).  We prioritize data diversity with respect to question styles, domain selection, and input lengths, as detailed in Table~\ref{tab:datasets}.
% Moreover, we place more emphasis on diversity in terms of question style, domain selection, and input length (Table~\ref{tab:datasets}).

In addition, the development of suitable evaluation metrics for LCLMs on open-ended tasks where multiple outputs are acceptable is crucial, yet challenging. In this work, we study the limitations of traditional metrics based on lexical matching. We demonstrate that these metrics often fail to correlate with human evaluation results. Our further experiments suggest that LLM judges~\citep{alpaca_eval, zheng2023judging} provide superior accuracy in the evaluation of open-ended tasks. \S\ref{sec:metric} explains how we set a short-context LLM judge in a long-context evalution setting. 
Considering the influence of generation length on performance and in order to avoid drawing misleading conclusions, we propose the Length-Instruction-Enhanced (LIE) evaluation technique for all reference-based metrics, including those employing an LLM judger. The empirical results demonstrate a substantial improvement brought by LIE evaluation in the Kendall-Tau correlation coefficient ($\tau$) with human judgments (Figure~\ref{fig:cor}), for all automatic metrics.

We also conducted a comprehensive study with 16 different LLMs (\S\ref{sec:baselines}) in L-Eval. Some of our key findings are summarized below:
(1) There is still a significant gap between open-source LCLMs and commercial models, for both closed-ended tasks (Table~\ref{table:acc_exam}) and open-ended tasks evaluated by LLMs and human (Table~\ref{tab:llm_eval},~\ref{tab:human_eval}). However, this gap is not accurately reflected by n-gram metrics.
(2) While current efforts on open-source LCLMs improve performance on closed-ended tasks, they significantly fall short on open-ended tasks. This is largely due to the models' misunderstanding of instructions as the input context length increases.
(3) Experiments on GPT-3.5-Turbo with both dense and sparse retrievers show that end-to-end full-context models outperform traditional retrieval-based systems.
(4) Training-free scaled positional embeddings can enhance the retrieval capability of LLMs over longer input, while it may adversely affect their reasoning ability.
% \begin{itemize}[leftmargin=12pt]
%     \item There is still a significant gap between open-source LCLMs and commercial models, for both closed-ended tasks (Table~\ref{table:acc_exam}) and open-ended tasks evaluated by LLM and human (Table~\ref{tab:llm_eval},~\ref{tab:human_eval}). Meanwhile, n-gram metrics can not truly reflect this gap.
    
%     \item Current efforts on open-source LCLMs can lead to better performance in closed-ended tasks but fail short in open-ended tasks due to misunderstanding instructions with input context length increase.
%     \item Experiments on GPT-3.5-Turbo show that retrieval-based models perform less effectively than full-context models.
%     \item Training-free scaled positional embedding faces a challenge in balancing retrieval ability and reasoning ability.
% \end{itemize}

        % When the input tokens increase, open-source LCLMs often misunderstand instructions from the open-ended tasks, but they can follow the instructions from closed-ended tasks, especially in multiple choice settings.
    % \item In a comparison between LCLMs and retrieval-based methods~\citep{robertson2009probabilistic, ram2023incontext}, we observe that LCLMs encoding full context of the document, \lpk{todo: rewrite: despite the risk of losing resolution}, generally achieve better performance than \lpk{todo: rewrite: pairing a retrieval model with a 4k model}. \sansa{didn't mention this in main results. Determine whether to mention it}\lpk{sansa can you help to rewrite these 2 todos :P}
More interesting conclusions can be found in \S\ref{sec:main_results} and \S\ref{sec:analysis_app}. We hope \textit{L-Eval} and our findings contribute to a deeper understanding of current LCLM research and the further development of models and evaluation metrics.

% Figure~\ref{fig:overall} demonstrates the rankings of LLMs on L-Eval and we also share more insightful findings in Section \S\ref{sec:main_results}.

% 放到 分析Retrieval-based methods also get impressive results in the long lectures full of terms and definitions but they are usually limited to both the input and instructions style. For example, if the long sequence input is composed of complex in-context examples with CoT, retrieved passages will dramatically decrease the performance from 84 to 23.
\section{Related Work}
\subsection{Long Context Language Models}\label{sec:lclms}
Feeding long context leads to bottlenecks in language model training and inference due to computational resources. Some community efforts focus on developing 
 {efficient attention} mechanisms to build efficient language models~\citep{sun2023retentive,ding2023longnet,li2023incontext,fu2023hungry,peng2023rwkv}. 
% They have made improvements to the attention mechanism by optimizing algorithms and IO operations, resulting in a significant reduction in both time and space requirements when handling long contexts.
In addition to optimizing the attention mechanism, some works~\citep{bulatov2023scaling,dai-etal-2019-transformer, mohtashami2023landmark} focus on {chunking the input} to model both the current text in the chunk and the previous context states, effectively extending the length of context processing.
Besides the efficiency challenge, the {scalability of positional embedding} is also crucial. ALiBi~\citep{press2022train}, and \textsc{xPos}~\citep{sun2022lengthextrapolatable} emphasize the significance of local context to enhance the language model's ability to perform extrapolation.
Moreover, position interpolation (PI)~\citep{chen2023extending} and NTK-aware~\citep{fixedNTK, dynamicNTK} are the most popular approaches based on RoPE~\citep{su2022roformer} to efficiently and effectively extend the context length. However, these works mainly validated their methods with perplexity (PPL)~\citep{sun2021long, fixedNTK}, and there has not been systematic validation on practical tasks.

% Another approach to handle long contexts is \textbf{retrieval-based} language models~\citep{shi2023replug,ram2023incontext,pmlr-v119-guu20a,izacard2022few,bertsch2023unlimiformer}. 
% Retrieval-based language models rely on retrieving pre-existing text segments or passages as their basis for generating responses. However, this also limits its application to tasks that require long document reasoning, for example, inquiring about the discussed topics in any location of the previous chat history or effectively merging information from various parts of the document.
% In these tasks, retrieval methods fall short of fully comprehending the full text and grasping intricate contextual relationships.

\subsection{Long Sequences Benchmarks}
\cite{tay2020long} introduce the Long Range Arena (LRA), a benchmark encompassing five distinct classification tasks. CAB~\citep{zhang2023cab} is another benchmark for different efficient attention designs by comparing both efficiency and accuracy. In language domain, 
previous work on LCLMs tends to report PPL to evaluate language models~\citep{su2022roformer, peng2023yarn} on longer context. However, PPL may not usually correlate with the actual performance~\citep{sun2021long}.  ZeroScrolls~\citep{shaham2022scrolls, shaham2023zeroscrolls} and LongBench~\citep{bai2023longbench} are concurrent long context evaluation suites. L-Eval differs from them in 3 aspects: (1) Manually selected samples. Testing samples are automatically filtered by their benchmarks, while those for L-Eval are manually filtered. (2) Standardized metrics.  We are the first to investigate the correlations between traditional lexical metrics and recently proposed LLM metrics with human judgment on Long context settings. L-Eval no longer mainly relies on N-gram metrics.  (3) More closed-ended tasks. Due to fairness issues in open-ended tasks. L-Eval has more closed-ended tasks reflecting unbiased results.

% As for the language domain, the most related work is SCROLLS~\citep{shaham2022scrolls} which collects and unifies the input format from 7 popular long document datasets. ZeroSCROLLS~\citep{shaham2023zeroscrolls} is their Zero-Shot setting with additional 4 datasets: SQuALITY~\citep{wang-etal-2022-squality}, MuSiQue~\citep{trivedi2022musique}, SpaceDigest~\citep{angelidis-etal-2021-extractive}, BookSumSort~\citep{kryscinski-etal-2022-booksum}.  Our work differs from SCROLLS in 2 aspects:
% (a (b) The conclusions drawn can vary significantly based on the chosen evaluation approaches and baselines. For open-ended tasks, SCROLLS relies on n-gram matching metrics which have been proven to have some inherent merits. For example, as we can see from Table~\ref{table:rouge2}, many open-source models have even exceeded turbo on summarization tasks according to ROUGE. However, as we can see from the GPT4 evaluation results~\ref{tab:llm_eval}, there is still a considerable gap. 
% Our results have led us to somewhat divergent findings: While SCROLLS suggests that LLMs are struggling in summarization tasks, our observations have demonstrated that LLMs perform quite well in these tasks.  The conclusion in SCROLLS may largely stem from the mismatch in length between LLM-generated summaries and the reference, significantly impacting n-gram metrics. Lastly, it's worth noting that L-Eval provides a broader analysis of open-source instruction-following LCLMs, while SCROLLS' assessments focus solely on Flan-T5~\citep{chung2022scaling}.

% \section{Data Collection and Annotation for L-Eval}

\section{Towards High-Quality and Diverse Long Context DataSets}
\label{sec:data}
In this section, we highlight some key procedures in L-Eval data construction.
Concretely, we show the annotation, re-annotation, and manual filtering pipeline and the statistics of L-Eval.  Please refer to Appendix~\ref{sec:data-appendix} for the complete annotation details and examples.

\subsection{Data Annotation from Scratch}\label{sec:scratch}
There are 4 datasets annotated from scratch in L-Eval: Coursera, SFcition, CodeU, and LongFQA. The original resources are videos from Coursera, previous open-source datasets, source code from famous Python libraries, and public earning call transcripts, respectively.

\vspace{-0.7em}
\paragraph{Coursera} This dataset originates from the Coursera website.\footnote{\url{https://coursera.org/}} To reduce the difficulty of annotation, we choose four public courses related to big data and machine learning (\S\ref{sec:coursera}).
The input long document is the subtitles of the videos. Questions and the ground truth answers are labeled by the authors. The instruction style of Coursera takes the format of multiple choice. In order to increase the difficulty of the task, we have set \textbf{multiple correct options}. To the best of our knowledge, this is the first multi-choice dataset with multiple correct answers and it is more challenging than single-option questions (Table~\ref{table:acc_exam}).

\vspace{-0.7em}
\paragraph{SFcition} 
We annotate this sub-task to test the loyalty of the LCLM to the input context. We argue that in LCLMs, contextual knowledge (stored in long input) is more crucial than parametric knowledge (gained during pretraining). Practically, many long documents are private and can never be seen during pretraining. LLMs should follow the contextual knowledge instead of parametric knowledge in long context settings. To simulate this scenario, we annotate a science fiction dataset consisting of True or False questions. Most of the answers to these questions contradict real-world principles and do not comply with actual physical laws (\S\ref{sec:sfiction}). We find that Turbo-16k struggles on this task, which tends to answer questions relying on parametric knowledge (Table~\ref{table:acc_exam}).

\vspace{-0.7em}
\paragraph{CodeU} As a code understanding dataset, it requires LLM to infer the output of a lengthy Python program. We mainly use source code from Numpy\footnote{\url{https://github.com/numpy/numpy}} and construct a string processing codebase. To prevent LLMs from answering the question based on their parametric knowledge, we replace the original function name. LLMs should first locate where the function is called and determine which functions are invoked. CodeU is the most challenging task in L-Eval (\S\ref{sec:codeU}).

\vspace{-0.7em}
\paragraph{LongFQA} We also notice that there is a lack of long context question answering datasets in the finance domain and we annotate the QA pairs based on public earning call transcripts from the \textit{Investor Relations} section of 6 company websites.  Please refer to \S\ref{sec:longfqa} for details.

\subsection{Data Re-annotation from Public Datasets}\label{sec:re-anno}
We re-annotate 5 publicly available datasets in L-Eval. 
\textbf{GSM(16-shot)} is derived from 100-grade school math problems in the GSM8k dataset~\citep{cobbe2021training}. If the LCLM maintain its reasoning ablilty on longer context, ultizing more high-quality examples will a positive effect on solving math problems~\citep{li2023incontext}. We construct 16 in-context examples with lengthy Chain-of-Thought where 8 examples come from  \textit{chain-of-thought-hub}\footnote{\url{https://github.com/FranxYao/chain-of-thought-hub}} and 8 examples are constructed by us.  We experiment with the newly constructed examples and the accuracy of Turbo-16k-0613 rises from 79 (8-shot) to 84 (16-shot).  

We inject come new synthesis instructions to test global context modeling into \textbf{QuALITY}~\citep{pang2022quality}, such as \textit{``What can we infer from the longest sentence in this story?''} and \textit{``How many words are there in the story?''}. Given that these types of questions may rarely occur in real-world conversations, their proportion in L-Eval is extremely small. 
The \textbf{Openreview} dataset contains papers collected from \url{openreview.net}. We ask the model to (1) write an Abstract
section, (2) summarize the related work, and (3) finally give feedback including valuable suggestions and 
some questions for the authors. We select the paper with high-quality related work sections and helpful reviews written by human reviewers to form this test set.\footnote{Ethic statement:
we discourage reviewers from using large models for reviews. Our goal is to assist authors in further improving their papers.} Next, we use \textbf{SPACE}~\citep{angelidis-etal-2021-extractive} to test the aspect-based review summarization task, and the instructions for the dataset are annotated by us. We adopt diverse instructions to prevent overfitting.

\begin{wrapfigure}{r}{0.4\textwidth}
  \centering
  \vspace{-4mm}
  \includegraphics[width=0.4\textwidth]{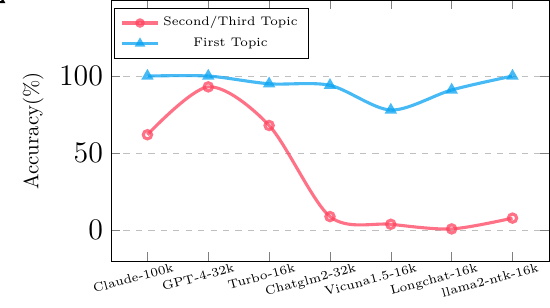}
  \caption{Test Accuracy (\%) of different models with retrieving the first topic and retrieving the second/third topic.}
  \label{fig:topic}
  \vspace{-5mm}
\end{wrapfigure}
Previous work~\citep{longchat2023,liu2023lost} has used retrieval tasks to test the ability of modeling long context dependency via retrieving something over lengthy context. L-Eval includes a popular first topic retrieval task \textbf{TopicRet}~\citep{longchat2023}, formatted as: ``\textit{[topic-1] Chat History [instruction]} ''. However, as we can see from Figure~\ref{fig:topic}, retrieving the first topic is too easy to distinguish the ability of different models. However, the task of retrieving the second and the third topics presents a significantly higher level of challenge. It is observed that nearly all open-source models struggle in  task.  So we enhance the task with second/third topic retrieval.

\subsection{Data Filtering and Correction}\label{sec:postprocess}
The remaining 12 tasks originates from existing datasets following previous evaluation suites~\citep{zhang2023cab}.  However, L-Eval involves more human labor after data collection because we find the annotation quality of previous long sequence datasets fluctuates severely and there are many unanswerable questions that are unrelated to the context. These mistakes can hardly be corrected using the automatic preprocessing scripts in previous works. In L-Eval, all samples are manually filtered and corrected after data collection. Specifically, we use Claude-100k as our assistant to filter mistaken QAs and unanswerable questions. First, we input the lengthy document into Claude and request it to provide the answer and offer an explanation. If Claude produces an answer greatly mismatching the ground truth or states that we cannot deduce the answer from the context, we will either perform re-annotation or simply remove them. 

\begin{table}[t]
\vspace{-1em}
\centering  
\caption{This table presents the statistics of the L-Eval suite where \textbf{Question-style} indicates the type of task or the style of instruction in the dataset, \textbf{\#Doc} refers to the number of long documents, and \textbf{\#Instr} denotes the number of instructions provided for each long input. \textbf{Avg/Max len} signifies the average/maximum length of the document inputs. We tokenize the raw text with Llama2 tokenizer and report the number of tokens.
}
\vspace{-0.5em}
\resizebox{\textwidth}{!}{
\begin{tabular}{lllrrrr}
\toprule
\bf Dataset &\bf Question-style &\bf Domain &\bf Avg len & \bf Max len &\bf \#Instr  &\bf \#Doc \\
\midrule
\rowcolor{mypink!50}
\multicolumn{7}{c}{\textbf{\textit{Closed - Ended \, Tasks}}} \\
\midrule 

TOEFL & Multiple choice & English test & 3,907 & 4,171  &  269 & 15\\
GSM(16-shot)$^\dag$ & Solving math problems & In-context examples & 5,557  & 5,638 & 100 & 100  \\
QuALITY~\cite{}$^\dag$ & Multiple choice & Gutenberg & 7,169  & 8,560  & 202 & 15 \\
Coursera$^*$  & Multiple choice  & Advanced courses & 9,075 & 17,185 & 172  & 15 \\
TopicRet$^\dag$  & Retriving topics & Conversation  &  12,506 & 15,916  & 150 & 50 \\
SFcition$^*$ & True or False Questions & Scientific fictions & 16,381 & 26,918 & 64 & 7 \\
CodeU$^*$ & Deducing program outputs & Python Codebase  &  31,575 & 36,509 & 90 & 90 \\

\midrule
\rowcolor{mypink!50}
\multicolumn{7}{c}{\textbf{\textit{Open - Ended \, Tasks}}} \\
\midrule
MultiDoc2Dial & Goal-oriented dialogues & Grounded documents & 3,905 & 7888  & 136 & 20 \\
Qasper & QA on papers & NLP papers & 5,019 & 6,547  & 160 & 20 \\
LongFQA$^*$  & QA on earning call & Finance  & 6,032 & 7824  & 52 & 6 \\
NQ & QA from Google Search & Wikipedia & 23,698 & 47,726 & 104 & 20\\
CUAD & Extracting key information & Law & 30,966 & 68,625  & 130 & 20 \\
NarrativeQA & QA on narratives  & Gutenberg  & 62,335  & 210,541 & 182 & 20 \\
\midrule
Multi-News & Multi-doc Summarization & Multiple News articles  & 7,320& 19,278  & 11 &11 \\
GovReport & Single-doc Summarization & Government reports  & 7,495 & 27,128  &13 & 13 \\
BigPatent & Single-doc Summarization & Lengthy patents  & 7,718  & 12,867 & 13 & 13 \\
SummScreen & Transcripts Summarization & TV series transcripts  & 10,688 & 14,544  & 13 & 13 \\
Openreview$^\dag$ & Paper writing \& reviewing & Papers from Openreview  & 11,170 & 33,303 & 60 & 20 \\
QMSum &  Query-based summarization & Meeting transcripts  & 16,692 & 33,310 & 156 & 20 \\
SPACE$^\dag$ & Aspect-based summarization & Reviews on Hotels  &  19,978 & 22,158 & 120 & 20 \\
\bottomrule
\end{tabular}
}

\label{tab:datasets}
\end{table}

\subsection{Statistics}\label{sec:stat}
The statistics of L-Eval are shown in Table~\ref{tab:datasets}.
The L-Eval contains various question styles such as multiple choice questions (TOFEL~\citep{tseng2016towards}, QuALITY, Coursera), true or false questions (SFiction), math problems (GSM), code understanding (CodeU), goal-oriented dialogues (MultiDoc2Dial~\citep{Feng_2021}), extractive QA (CUAD~\citep{hendrycks2021cuad}, NQ~\citep{kwiatkowski-etal-2019-natural}), abstractive QA (LongFQA, NarrativeQA~\citep{kočiský2017narrativeqa}, Qasper~\citep{dasigi2021dataset}), single document summarization (GovReport~\citep{huang-etal-2021-efficient}, BigPatent~\citep{sharma-etal-2019-bigpatent}, SummScreen~\citep{chen2022summscreen}, QMSum~\citep{zhong2021qmsum}), multi-document summarization (Multi-News~\citep{fabbri2019multinews}, SPACE~\citep{angelidis-etal-2021-extractive}), research writing (Openreview) and so on. The long documents in L-Eval across many domains such as law, finance, academic papers, lectures, lengthy conversations, news, famous Python codebase, long-form novels, and meetings. The average input length in L-Eval ranges from 4k to 60k. The maximum sample in L-Eval contains nearly 200k tokens. This diversity represents real-world scenarios where different tasks may require different lengths of context and instructions. The length of reference in L-Eval also varies significantly across tasks.
% which is indicative of the different response formats expected in these tasks. Some tasks require short answers or selections, while others involve lengthier summaries or comprehensive responses.

\section{Towards Standardized Long Context Evaluation Metrics}
\label{sec:metric}
In this section, we present various evaluation metrics for text generation, including exam evaluation for close-ended tasks and different levels of open-ended evaluation, most of which are reference-based metrics.
We also conduct experiments to study the correlation between automated metrics and human scoring. 

\vspace{-0.7em}
\paragraph{Exam evaluation} This is designed for closed-ended tasks, i.e., multiple-choice questions. The evaluation metric used for these tasks follows the exact match format (accuracy \%), similar to grading exam papers. Each question's score is calculated as 100 divided by the number of questions. 

\vspace{-0.7em}
\paragraph{Human evaluation} This is the most accurate evaluation for open-ended tasks. Despite that some works show GPT-4 can be coherent with human judgment, LLMs cannot replace human evaluation. We engage human evaluators to score the outputs on a scale of 1 to 5, which signifies from poor output to excellent output.
To save human laboratories, we propose a subset used for the human evaluation which has 12 long documents with 85 open-ended questions (\textbf{85-question subset}). 

\vspace{-0.7em}
\paragraph{Large language model judges for evaluating LCLMs}
In short context settings,  evaluation using LLMs is the most accurate metric for automatically evaluating models on open-ended tasks~\citep{zheng2023judging,alpaca_eval, dubois2023alpacafarm}. These works assume the LLM evaluator is a ``super model'', but this assumption does not hold in long context settings because it's impossible to feed the entire lengthy inputs into LLMs like GPT-4. Unlike short context evaluation, GPT-4 is unable to infer the ground truth answer itself. Consequently, evaluation results mainly depend on the reference answer and user questions. In L-Eval, we take the pair-wise battle format and we select Turbo-16k-0613 as the base model and report the \textit{win-rate vs. Turbo-16k-0613 \%} which means how many samples can beat Turbo-16k. We study two LLM judges: GPT-4 and GPT-3.5 in the experiment section.
LLM evaluators have been reported to favor more detailed and lengthy answers~\citep{zheng2023judging}. This bias becomes more pronounced in long context settings as the invisible input makes it difficult for the judge to accurately determine the correctness of specific details and information. Therefore, the judgment model must bear in mind that details not corroborated by the reference answers should not be considered beneficial. We enhance the judgment prompt with: \textit{Additional details or information that are not mentioned in the reference answer cannot be considered as advantages and do not let them sway your judgment.} If you only want to evaluate a portion of the tasks in L-Eval, we recommend using LLM judges.
% \begin{example}{}{example}
% \textit{\footnotesize Please act as an impartial judge and evaluate the quality of the responses provided by two AI assistants to the user question user's query that was generated from the context of a long document.  You will be given a reference answer written by human, Assistant A's answer, and Assistant B's answer. Your job is to evaluate which assistant's answer is better. Begin your evaluation by comparing both assistants' answers with the reference answer. Additional details or information that are not mentioned in the reference answer cannot be considered as advantages and do not let them sway your judgment...
% }
% \end{example}
Verifying the 1000+ open-ended questions via GPT-4 is unaffordable.\footnote{Testing the 4 datasets in Table~\ref{tab:length_bias} needs about \$100!} Thus we manually split a subset for GPT-4 evaluation consisting of 17 diverse long documents with 96 open-ended questions (\textbf{96-question subset}).\footnote{Evaluating outputs from the 96-question subset with GPT-4 only needs about \$5.}

\vspace{-0.7em}
\paragraph{N-gram matching evaluation}
Considering that assessing all tasks is still expensive for human/LLM evaluators, L-Eval also takes into account n-gram metrics. 
N-gram metrics like ROUGE-L (R-L) and F-1 score are widely used in traditional datasets and they are also widely adopted in the text generation benchmarks via performing lexical matching. It is worth noting that n-gram matching metrics are very sensitive to the length of the ground truth, exhibiting a length bias. The related analysis is in the following \S\ref{sec:length-instruction}.

\subsection{Length Instruction Enhanced Long Context Evaluation}
\label{sec:length-instruction}

\begin{wrapfigure}{r}{0.4\textwidth}
  \centering
  \vspace{-6mm}
  \includegraphics[width=0.4\textwidth]{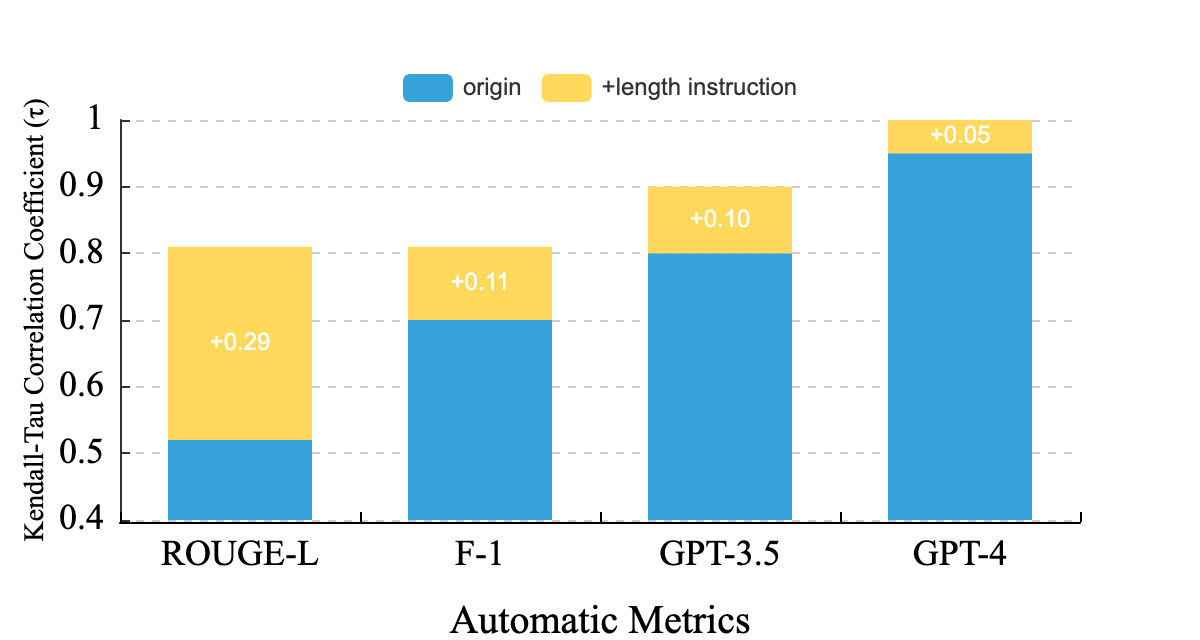}
  \caption{Kendall-Tau correlation coefficient of different automatic metrics with the average human score.}
  \label{fig:cor}
  \vspace{-5mm}
\end{wrapfigure}
In preliminary experiments, we find that LLMs tend to generate very long responses bringing obstacles for the reference-based evaluation (see $\Delta$\textbf{L} Table~\ref{tab:length_bias}). This length bias results in a significant influence on the n-gram metrics. For instance, Claude-100k only achieves a 9.84 F-1 score due to undesired output length.

In L-Eval, we argue that long context language models should further focus on more accurate content rather than accurate length. Practically, issues about undesired generation length can be easily solved by prompting the model.  We first adopt \textbf{Length-Instruction-Enhanced} (LIE) evaluation in LLMs evaluation benchmarks which is simple but effective in overcoming the length bias, i.e., the number of words of ground truth is directly exposure to LCLMs. 
LIE evaluation in this work is implemented by injecting the model with the desired length into the original instruction (e.g., [Origin Instruction]: \textit{Please summarize the opinions of the professor}. [Length Instruction]: \textit{We need a 50-word summary}, where 50 is the number of words in the reference answer). The results of Claude-100k in Table~\ref{tab:length_bias} demonstrate a substantial improvement in terms of the F-1 score: there is a near \textbf{50}-point gap depending on whether or not the model generates with the expected length.

\begin{figure}[t]
\vspace{-1em}
    \centering
    \includegraphics[width=0.85\textwidth]{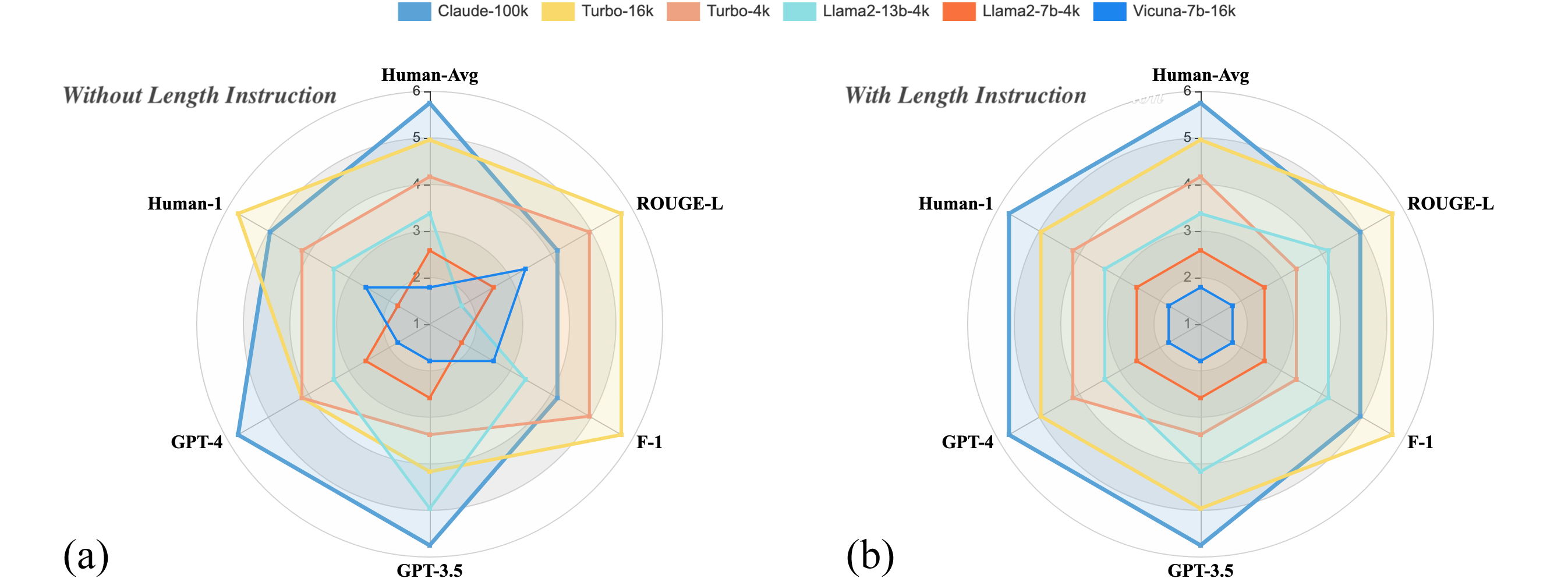}
    \vspace{-0.5em}
    \caption{The ranking of six models under various evaluation metrics (Human-avg, Human-1, GPT-4, GPT-3.5, R-L, and F-1) with or without length instruction. {Human-avg} represents the average score from human evaluation, and {Human-1} signifies the score given by the first human annotator.}
    \label{fig:radar}
\end{figure}

\vspace{-0.7em}
\paragraph{Experimental validation} To validate the LIE evaluation, we then conduct a human evaluation on the 85-questions subset. We have 3 annotators to verify 7 models and calculate the Kendall-Tau correlation coefficient ($\tau$) between these metrics and the average human score. The main results are shown in Figure~\ref{fig:cor} (Blue bar) and experimental settings are in \S\ref{sec:appendix:human-eval}. Results indicate that all these automatic metrics (except GPT-4) \textbf{fail to correlate} to human judgment. Compared with N-gram metrics, LLM judges are more accurate and robust to output length.
As we can see from Figure~\ref{fig:cor}, the improvements brought by length instruction are marked with yellow, and after adding the length instructions, $\tau$ has been improved from 0.5 to 0.8 for ROUGE-L and $\tau$ of GPT-4 evaluator has even reached to 1. In Figure~\ref{fig:radar}, we convert the score to rankings (the best one is 5 and the worst is 1) and show the score of 6 models evaluated with 6 different evaluation systems. Figure~\ref{fig:radar} (a) shows the results given by metrics without length instruction. These hexagons are often distorted because these metrics usually cannot achieve good correlation. When comparing the models enhanced with length instruction in (b), it is observed that the hexagons become more regular.

\begin{table}[t]
\vspace{-1em}
\centering
\caption{Results on 2 open-ended summarization and 2 abstractive QA tasks.  \textbf{GPT-4} means the win-rate with Turbo-16k using GPT-4 as the judge.  \textbf{$\Delta$L} means the difference of generated answer length with ground truth length. The best results are underlined. Results in red mean decoding in a desired length makes a big difference in performance.}
\vspace{-0.5em}

\resizebox{\textwidth}{!}{
\begin{tabular}{l|ccc|ccc|ccc|ccc}
\toprule
\multirow{2}{*}{\textbf{Model}} & \multicolumn{3}{c|}{\textbf{SPACE}} & \multicolumn{3}{c|}{\textbf{QMSum}} & \multicolumn{3}{c|}{\textbf{NQ}} & \multicolumn{3}{c}{\textbf{NrtvQA}} \\
\cmidrule(lr){2-4} \cmidrule(lr){5-7} \cmidrule(lr){8-10} \cmidrule(lr){11-13}
&\textbf{R-L} &  \textbf{GPT-4} &  \textbf{$\Delta$L} & \textbf{R-L} & \textbf{GPT-4} & \textbf{$\Delta$L}  & \textbf{F-1}  & \textbf{GPT-4} & \textbf{$\Delta$L}  & \textbf{F-1} & \textbf{GPT-4}  & \textbf{$\Delta$L}  \\
\midrule

\rowcolor{mypink!60}
Claude-100k & 15.43 & \color{red}{45.65} & \bf165 & 14.04 & 58.77 & \bf183 & \color{red}{9.84} & \underline{56.19} & \bf135 & \color{red}{10.39} & \underline{68.96} & \bf127 \\
 \quad + Length Instruction &18.61 & \color{red}{\underline{61.40}} & 27 & 18.13 & \underline{58.89} & 22 & \color{red}{\underline{57.76}} & {51.00} & 1 & \color{red}{\underline{19.09}} & 57.77 & 0 \\

\rowcolor{mypink!60}
Chatglm2-32k & 17.56 & 24.13 & -23 & 20.06 & 38.84 & \bf287 & 31.45 & 33.71 & 3 & 12.24 & 34.67 & 74 \\
 \quad + Length Instruction & 16.61 & 17.11 & 11 & \underline{20.83} & 33.75 & 9 & 37.94 & 33.71 & -1 & 14.00 & 34.52 & -2 \\

\rowcolor{mypink!60}
Longchat-7b-16k  & 15.10 & \color{red}{15.61} & \bf120 & 9.31 & 25.56 & 40 & {8.83} & 32.33 & \bf105 & 8.36 & 31.80 & 83 \\
 \quad + Length Instruction& 17.06 & \color{red}{36.23} & -3 & 13.21 & 30.20 & 70 & {20.21} & 35.00 & 37 &15.17&43.38&40\\

\rowcolor{mypink!60}
Llama2-13b-chat & 16.83&32.46&\bf102&14.72&30.79& \bf116 &\color{red}{8.29} &38.99&90&7.20 & \color{red}{30.69} & \bf130 \\
 \quad + Length Instruction & \underline{19.23} & 43.15 &-7 &19.65 &34.82 &-1&\color{red}{35.43} &41.07 & 6 & 13.48 & \color{red}{45.07} & 14 \\

% Turbo-16k (Anchor) &18.67 & 50.00 & -19 & 19.31 & 50.00 & 26 & 45.90 & 50.00 & 15 & 18.20 & 50.00 & 2 \\
\bottomrule
\end{tabular}
}
\label{tab:length_bias}
\end{table}

\section{Benchmarking LLMs with L-Eval}
In this section, we list our 16 baseline models and the results on both open-ended and closed-ended tasks. Generally, there are considerable gaps between open-source models and commercial models. A detailed description of baseline models can be found in \S\ref{sec:baselines_appendix}.  The prompt templates for each task are available in \S\ref{sec:data-appendix}. We run all the experiments using FlashAttention~\citep{NEURIPS2022_67d57c32} on a single NVIDIA A800 GPU. The document input is truncated from the right.
% \paragraph{There are considerable gaps between open-source models and commercial models} The main results on open-ended tasks are shown in Table~\ref{tab:llm_eval}, the commercial models GPT3.5-16k, GPT4-32k, and Claude-100k are more advanced compared to the open-source models. Although Claude-100k falls behind GPT-4 on many short context benchmarks, it generally performs better than GPT-4 and achieves the best results.  For closed-ended tasks,  as we can see in Table~\ref{table:acc_exam}, the advantage remains more evident. However, as for those open-ended generation tasks especially summarization tasks, the difference between commercial models and open-source models is not significant as closed-ended tasks via indicating the length of the ground truth. We assume that the summarization tasks require less reasoning ability from models and instead rely more on basic language generation capacity, which can be easily accomplished by the listed LLMs. Closed-ended tasks usually need the model to generate the output following a specific format. Another reason is that open-source models struggle to effectively and fully obey the instructions leading to the undesired output extracted from all generated tokens.

\vspace{-0.5em}
\subsection{Baselines}\label{sec:baselines}
\vspace{-0.3em}
\paragraph{Commercial Models}
(1) {Claude-100k} developed by Anthropic, (2) GPT-4-32k, OpenAI's most powerful long context model, (3)
{Turbo-4k-0613} and (4) Turbo-16k-0613 is the snapshot of {GPT-3.5} from June 13th 2023 which can handle up to 4k/16k input tokens.
\vspace{-0.7em}
\paragraph{Open-source Models}
(5) {Llama1}~\citep{touvron2023llama}, a widely used open-source model developed by Meta AI with a 2k pre-training length, (6) {Vicuna1.3}~\citep{vicuna2023}, tuned on shareGPT based on Llama1, (7) Longchat-16k, the long context version of Vicuna1.3 using PI, (8) Llama2, the next version of Llama with 4k pre-training context, (9) Llama2-chat, a finetuned version for dialogue usage, (10) Llama2-NTK, extending the context length of Llama2-chat with NTK-aware RoPE, (11) Vicuna1.5-16k~\citep{zheng2023judging}, the long context version of Llama2 using PI \& ShareGPT (12) Longchat1.5-32k, the 32k context version of Llama2 using PI \& ShareGPT.
(13) {Chatglm2-8k}, the second version of the Chatglm~\citep{du2022glm}, (14) Chatglm2-32k, the 32k context length version,  (15)\ {XGen-8k-inst}~\citep{XGen}, an 8k context models developed by salesforce  (16) {MPT-7B-StoryWriter-65k}, based on MPT-7B and ALiBi with a context length of 65k tokens on a subset of Books3 dataset.
\vspace{-0.7em}
\paragraph{Retriever}  We implement the dense retriever with the OpenAI AdaEmbedding as the dense retriever and BM25 as the sparse retriever to extract 4 pieces of most related 1k-chunked documents, which are further provided as the context to answer questions.
% The prompt used as follows:
% \begin{example}{}{example}
% \textit{\footnotesize Use the following pieces of context to answer the question at the end. If you don't know the answer, just say that you don't know, don't try to make up an answer. \{context\} Question: \{question\}
% }
% \end{example}

\begin{table}[t!]
\vspace{-1.5em}
\centering
\setlength{\tabcolsep}{1.0mm}
\caption{Exam evaluation results on  \textbf{closed-ended tasks} for current LCLMs. \textbf{Ret.} indicates whether we use retrieve-based algorithms for the base model. \textbf{Tokens} denotes the maximum number of input tokens we feed into the model. {\color{red}{\tiny{$\downarrow$ / $\uparrow$}}} indicates a remarkable decrease/increase in performance, compared to using the original short context counterpart. {\color{blue} *} indicates the model is not further trained.}
\vspace{-0.5em}
\renewcommand\arraystretch{1.05}
% \begin{tabular}{@{}lcccccccccc@{}}
\resizebox{\textwidth}{!}{
\begin{tabular}{@{}lcccccccccc@{}}
% \begin{tabular}{lllrrrr}
\toprule
\textbf{Model} & \textbf{Ret.} & \textbf{Tokens} & \textbf{Coursera} & \textbf{GSM} & \textbf{QuALITY} & \textbf{TOEFL}  & \textbf{CodeU} & \textbf{SFiction} &  \textbf{Avg.} \\
\midrule
Claude1.3-100k & \xmark & \cellcolor{gray!70}100k & 60.03 & {88.00} & {73.76} & {83.64} & 17.77 & 72.65 & 65.97  \\
GPT-4-32k & \xmark & \cellcolor{gray!60}32k & \textbf{75.58} & \textbf{96.00} & \textbf{82.17} & \textbf{84.38} & \textbf{25.55} & \bf74.99 & \bf73.11 \\
Turbo-16k-0613 & \xmark & \cellcolor{gray!40}16k & 63.51 & {84.00} & 61.38 & 78.43 & 12.22 & 64.84 & 60.73 \\
AdaEmb-Turbo-4k-0613 & \cmark & \cellcolor{gray!15}4k & 61.77 & {23.00} & 58.91 & 76.95 & {6.66} &71.09 & 49.73 \\
BM25-Turbo-4k-0613 & \cmark & \cellcolor{gray!15}4k & 63.80 & {23.00} & 59.40 & 75.09 & {5.55} & 71.09  & 49.65 \\

\midrule
\rowcolor{mypink!50}
\multicolumn{10}{c}{\textit{Truncating input tokens to the pretraining context length}} \\
\midrule

Llama1-7b-2k (w/o SFT) & \xmark & \cellcolor{gray!5}2k & 13.37 & 7.00 & 21.78 & 30.85 & 1.11 & 35.15  & 19.22 \\
Vicuna1.3-7b-2k & \xmark & \cellcolor{gray!5}2k & 34.73 & 19.00 & 32.67 & 43.49 & 1.11 & 60.93 & 30.01 \\
% Longchat-7b-16k & \xmark & \cellcolor{gray!5}2k & 34.44 & 7.00 & 35.14 & 43.12 &  \bf5.55 & \bf64.05 & 31.54 \\

Llama2-7b-4k (w/o SFT) & \xmark & \cellcolor{gray!15}4k  & 20.05 & 2.00 & 28.71 &  24.53 & 0.00 & 40.62  & 19.31 \\
Llama2-7b-chat & \xmark & \cellcolor{gray!15}4k  & 29.21 & {19.00} & 37.62 & 51.67 & 1.11 & 60.15 & 33.12 \\
Llama2-13b-chat & \xmark & \cellcolor{gray!15}4k  & 35.75 & \textbf{39.00} & \textbf{42.57} & \textbf{60.96} & 1.11 & 54.68 & \bf39.01 \\

Chatglm2-6b-8k & \xmark & \cellcolor{gray!5}2k & \bf43.75 & {13.00} & {40.59} & {53.90} & 2.22 & 54.68 & 34.69  \\
XGen-7b-8k (2k-4k-8k) & \xmark & \cellcolor{gray!5}2k & 26.59 & 3.00 &  35.15  & 44.23 & 1.11 & 48.43 & 26.41 \\

\midrule
\rowcolor{mypink!50}
\multicolumn{10}{c}{\textit{Truncating input tokens to the further finetuning context length}} \\
\midrule
% longchat1.5-7b-32k & \xmark & \cellcolor{gray!60}32k  & 41.13 & 18.00 & 37.62 & 39.77 & -- & 29.33 \\
Chatglm2-6b-32k & \xmark & \cellcolor{gray!60}32k  & \textbf{47.81} & 27.00\color{red}{\tiny{$\uparrow$}} & 45.04 & 55.01 & 2.22 & 57.02 & 39.01\color{red}{\tiny{$\uparrow$}} \\
Longchat1.5-7b-32k & \xmark & \cellcolor{gray!60}32k  & 32.99 & 18.00 & 37.62 & 39.77 & 3.33 & 57.02 & 31.45\\

Longchat-7b-16k & \xmark & \cellcolor{gray!40}16k  & 29.74 & 10.00\color{red}{\tiny{$\downarrow$}}  & 33.66 & 47.95 & 3.33 & \bf64.84 & 31.58\color{red}\\ 
Vicuna1.5-7b-16k & \xmark & \cellcolor{gray!40}16k  & 38.66 & 19.00 & 39.60 & 55.39 & \bf5.55 & 60.15 & 36.39\color{red}{\tiny{$\uparrow$}} \\
Llama2-7b-NTK\color{blue}{*} & \xmark & \cellcolor{gray!40}16k  & 32.71 & 19.00 & 33.16 & 52.78 & 0.00 & \bf64.84 & 33.74 \\

Longchat-13b-16k & \xmark & \cellcolor{gray!40}16k & 31.39 & 15.00 & 40.59 & 55.39 &  2.22 & \bf64.84  & 34.90 \\
Vicuna1.5-13b-16k & \xmark & \cellcolor{gray!40}16k & 40.69 & 36.00 & \textbf{53.96}\color{red}{\tiny{$\uparrow$}} & \textbf{68.40}\color{red}{\tiny{$\uparrow$}} & 0.00 & 61.71 & \bf43.46\color{red}{\tiny{$\uparrow$}}\\
Llama2-13b-NTK\color{blue}{*} & \xmark & \cellcolor{gray!40}16k  & 36.48 & 11.00\color{red}{\tiny{$\downarrow$}} & 35.64  & 54.64&  1.11  & 63.28 & 33.69 \\
Llama2-13b-NTK(Dyn)\color{blue}{*} & \xmark & \cellcolor{gray!40}16k  & 30.08 & \textbf{43.00} & 41.58 & 64.31 & 1.11 & 35.15 & 35.87\\

% 8k baselines
% longchat-13b-16k & \xmark & \cellcolor{gray!25}8k & 30.23 & 10.00 & 37.12 & {55.76} & 2.22 & \bf64.84 & 33.36 \\
Chatglm2-6b-8k & \xmark & \cellcolor{gray!25}8k & 42.15 & {18.00} & {44.05} & {54.64} & 2.22 &54.68 & 35.95\\
XGen-7b-8k & \xmark & \cellcolor{gray!25}8k & 29.06 & 16.00 &  33.66  & 42.37 & 3.33 & 41.40 & 27.63\\
MPT-7b-65k & \xmark & \cellcolor{gray!25}8k & 25.23 & 8.00 & 25.24 & 17.84 & 0.00 & 39.06 & 19.22\\

\bottomrule
\end{tabular}
}

\label{table:acc_exam}
\end{table}

% separate tasks
\begin{table*}[t]
\vspace{-1.5em}
\caption{In comparing various models to Turbo-16k-0613 on \textbf{open-ended tasks}. We evaluate these models on the 96-question subset using GPT-4 and two subsets (85+96 questions) using GPT-3.5. We reduce the positional biases by swapping paired predictions, so the GPT-4 evaluator is used in 96$\times$2 evaluation rounds, while the GPT3.5 evaluator is used in 181$\times$2 rounds}
\vspace{-1.0em}
\center
\footnotesize
\renewcommand\arraystretch{0.97}
\resizebox{0.9\textwidth}{!}{
\tabcolsep 0.035 in
\begin{tabular}{lccccccccc}
\toprule
\multicolumn{1}{c}{\multirow{2}[1]{*}{\textbf{Model}}} &
\multicolumn{1}{c}{\multirow{2}[1]{*}{\textbf{Ret.}}} &
\multicolumn{1}{c}{\multirow{2}[1]{*}{\textbf{Tokens}}} &
\multicolumn{3}{c}{\textbf{GPT-4 }}&
\multicolumn{3}{c}{\color{darkgray}{\textbf{GPT-3.5}}} &
\multicolumn{1}{c}{\multirow{2}[1]{*}{\color{gray}{\textbf{R-L}}}}
\\
 & & & \textbf{wins} & \textbf{ties} & \textbf{win-rate \%}  &
\textbf{wins} & \textbf{ties} & \textbf{win-rate \%}  \\
\cmidrule(lr){1-1} \cmidrule(lr){2-2} \cmidrule(lr){3-3} \cmidrule(lr){4-6} \cmidrule(lr){7-9} \cmidrule(lr){10-10} 

Claude1.3-100k & \xmark & \cellcolor{gray!70}100k & \textbf{96} & 42 & \textbf{60.94}  & 189 & 34 & {\textbf{58.68}} & 28.22 \\
GPT-4-32k & \xmark & \cellcolor{gray!60}32k & 76 &  56 & 54.16  &  171 & 50 & {56.32} & \underline{36.18} \\
Turbo-16k-0613 & \xmark & \cellcolor{gray!15}4k & 0 & 192 & 50.00  & 0 & 362 & 50.00 & 28.61\\

Turbo-4k-0613 & \xmark & \cellcolor{gray!15}4k & 38 & 69 & 39.83\color{red}{\tiny{$\downarrow$}}  & 109 &  61 & {41.39} & 26.90\\
AdaEmb-Turbo-4k-0613 & \cmark & \cellcolor{gray!15}4k & 61 & 56 &  46.84 & 123 & 77  & 45.36 & 26.09 \\
BM25-Turbo-4k-0613 & \cmark & \cellcolor{gray!15}4k & 50 & 69 & 44.01 & 125 & 78 & {45.30} & 26.83\\

\midrule
\rowcolor{mypink!50}
\multicolumn{10}{c}{\textit{Truncating input tokens to the pretraining context length}} \\
\midrule

Vicuna1.3-7b-2k  & \xmark & \cellcolor{gray!5}2k & 29 & 55 & 29.42 & 97 & 42 & 34.91 & 16.17 \\
Longchat-7b-16k & \xmark & \cellcolor{gray!5}2k & 26 & 63 & 29.94  &  87 &  38 & {31.26} & 19.77\\

Llama2-7b-chat  & \xmark & \cellcolor{gray!15}4k & 48 & 58 & 40.10 & 127 & 44 & {42.45}  & \underline{24.25}\\
Llama2-13b-chat & \xmark & \cellcolor{gray!15}4k & \textbf{51} & 61 & \textbf{42.44}  & \textbf{143} & 49 & \textbf{47.85} & 24.07  \\
% Llama2-7b-chat  & \xmark & \cellcolor{gray!5}2k & 49 & 46 & 37.50\tiny{$\downarrow$}  & 105 &  51 & {37.39}\tiny{$\downarrow$}  & 23.05\\

\midrule
\rowcolor{mypink!50}
\multicolumn{10}{c}{\textit{Truncating input tokens to the further finetuning context length}} \\
\midrule

Chatglm2-6b-32k & \xmark & \cellcolor{gray!60}32k & 28 & 60 & 30.20 &  53 & 65 & {24.63} & \underline{22.04} \\
Longchat1.5-7b-32k & \xmark & \cellcolor{gray!60}32k & \bf38 & 53 & 33.59 &  136 & 37 & 44.91 & 21.21\\

Longchat-7b-16k & \xmark & \cellcolor{gray!40}16k & 36 & 56 & 33.68\color{red}{\tiny{$\uparrow$}} & 108 & 42 & {37.94} & 20.59\\
Vicuna1.5-7b-16k & \xmark & \cellcolor{gray!40}16k & 22 & 54 & 25.52\color{red}{\tiny{$\downarrow$}} & 102 & 52 & {37.86} & 18.05 \\
Llama2-7b-NTK\color{blue}{*} & \xmark & \cellcolor{gray!40}16k & 18 & 49 & 22.13 & 58 & 35 & {23.59} & 11.50 \\

Longchat-13b-16k & \xmark & \cellcolor{gray!40}16k & 36 & 59 & \textbf{34.11} & \bf128 & 24 & 40.11 & 18.98\\
Vicuna1.5-13b-16k & \xmark & \cellcolor{gray!40}16k & 36 & 59 & \textbf{34.11}\color{red}{\tiny{$\downarrow$}} &  116 &  43 & \bf{40.92} & 19.69 \\
Llama2-13b-NTK\color{blue}{*} & \xmark & \cellcolor{gray!40}16k & 31 & 52 & 29.68 & 91 & 44 & 34.55 &  15.63\\
Llama2-13b-NTK(Dyn)\color{blue}{*} & \xmark & \cellcolor{gray!40}16k &  23 & 48 & 24.47 & 55 & 64 & 26.60 & 11.62\\

Chatglm2-6b-8k & \xmark & \cellcolor{gray!25}8k & 18 & 64 & 26.04  &  86 & 54 & {32.84} & 18.19  \\
XGen-7b-8k & \xmark & \cellcolor{gray!25}8k & 24 & 62 & 28.64  & 89 & 72 & {36.02} & 20.51 \\
\bottomrule
\end{tabular}
}

\label{tab:llm_eval}
\end{table*}

\subsection{Main Results}\label{sec:main_results}
The performance of LCLMs on closed-ended tasks is shown Table~\ref{table:acc_exam}. As for open-ended tasks, we test the 96-question subset (Table~\ref{tab:llm_eval}) with GPT-4 evaluation. Results from n-gram metrics on all test sets and the rankings of LLMs can be found in \S\ref{sec:analysis_app}. From the main results, we have the following observations. 
GPT-4-32k clearly outperforms all other models by a very significant margin, establishing SOTA in L-Eval closed-ended tasks. 
% CodeU is the most challenging task, even the most powerful model only gets 25\% accuracy. \sansa{repeat. state this in data section}
There is still a near \textbf{20}-points gap between the best open-source 16k models and Turbo-16k.  As for open-ended tasks, since the input texts are generally longer and a global understanding of the context is required, Claude-100k, with the longest context length, surpasses all baseline models including GPT-4-32k. Although results of n-gram metrics indicate that open-source LCLMs have achieved performance close to GPT-Turbo on open-ended tasks, the evaluation outcomes from both LLM (Table~\ref{tab:llm_eval}) and human judges (Table~\ref{tab:human_eval}) reveal that there is still a significant gap between them. Moreover, retrieval-based methods based on Turbo-4k fall short in comparison to encoding the entire context (Turbo-16k), as certain tasks are difficult to address through simple retrieval.
% Figure~\ref{fig:overall} demonstrates the rankings of LLMs on L-Eval.

\vspace{-0.7em}
\paragraph{Fine-tuning longer offers benefits for closed-ended tasks but falls short in open-ended tasks} 
\begin{wrapfigure}{r}{0.4\textwidth}
  \centering
  \vspace{-5mm}
  \includegraphics[width=0.4\textwidth]{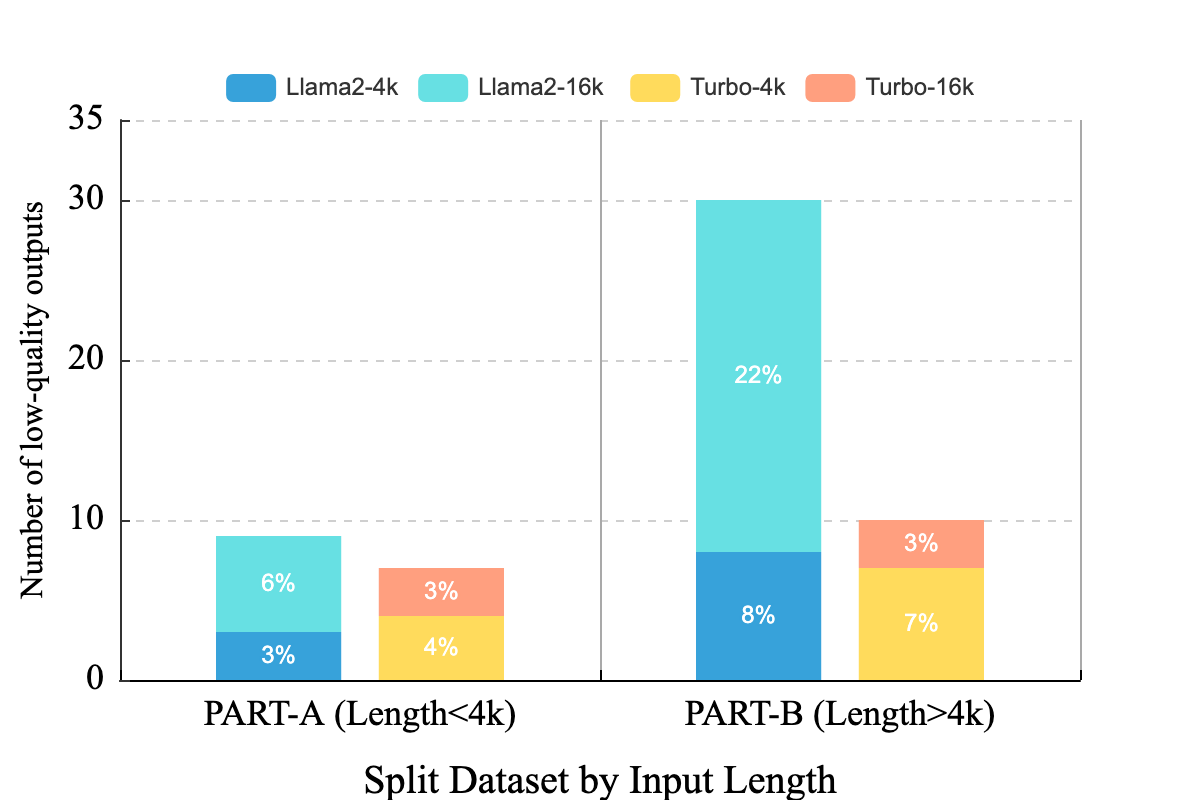}
  \caption{Number of invalid outputs from Llama2 and Turbo.}
  \label{fig:percent}
  \vspace{-3mm}
\end{wrapfigure}
In Table~\ref{table:acc_exam}, for open-source models using scaled positional embedding, Longchat and Vicuan1.5-16k obviously outperform their original version Vicuna-2k and Llama2-chat. The results suggest that further tuning on longer input from a model with short pretraining context length does benefit long context modeling. However, according to Table~\ref{tab:llm_eval}, unlike results on closed-ended tasks, the best model Vicuna1.5-13b-16k only wins Turbo-16k by 34\%, \textbf{8} points lower than its short version Llama2-13b. Llama2-13b-chat~\citep{touvron2023llama} is still the strongest open-source baseline, indicating that current LCLMs simply based on scaled position embedding may not be enough for these challenging open generation tasks. Based on our human evaluation, we find that although scaled position embedding techniques such as NTK~\citep{fixedNTK} or PI~\citep{sun2022lengthextrapolatable} effectively extend models' context length, the models tend to get lost when facing lengthy input tokens and are unable to follow the instruction. We classify these outputs as ``invalid outputs''. 
To investigate model performance on different context lengths, we split the 85-questions subset into 2 parts: PART-A contains samples with less than 4k tokens, and PART-B more than 4k tokens. We compare the number of invalid outputs from Llama2/Vicuna1.5-16k and Turbo/Turbo-16k in Figure~\ref{fig:percent}. Results show that the number of invalid outputs from Turbo-16k remains a very small amount on both PART-A and B while the invalid outputs from Llama2-16k dramatically increase on samples with longer input. Thus, LCLMs are less capable of following instructions on open-ended tasks for long contexts, compared with closed-ended tasks, such as multiple choice. A possible reason is that the pertaining or SFT corpus is highly likely to contain many training samples with similar question styles. This strongly enhances their instruction-following ability on closed-ended tasks. 

\vspace{-1em}
\paragraph{Performance on retrieval tasks contradicts reasoning tasks}
\begin{wrapfigure}{r}{0.35\textwidth}
  \centering
  \vspace{-1mm}
  \includegraphics[width=0.35\textwidth]{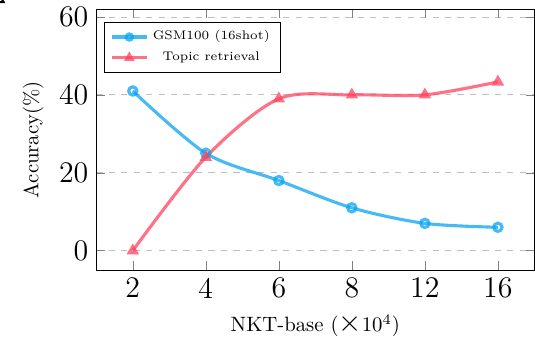}
  \caption{Test retrieval ability and reasoning ability with NTK base.}
  \label{fig:ret_reasoning}
  \vspace{-2mm}
\end{wrapfigure}
The most popular NTK-ware positional embedding methods increase the base 10,000 
% ($\theta_n={10000}^{-2n/d}$)
in the vanilla RoPE to implement extrapolation without further fine-tuning. However, we find that the performance on topic retrieval tasks does not match the reasoning capability over lengthy context. As can be seen from Figure~\ref{fig:ret_reasoning}, when we increase the base from 20,000 to 160,000, there is a continuous improvement on topic retrieval. However, performance on math reasoning tasks with lengthy examples exhibits a completely opposite trend, indicating that it is challenging for the model to maintain its reasoning abilities when increasing the base. In contrast, the performance on retrieval tasks seems to remain unaffected after the base reaches 60,000.

We have further analysis in \S\ref{sec:analysis_app}, including full results of n-grams metrics on open-ended tasks, the rankings of current LLMs, NTK-aware positional embedding and retrieval-based systems.

\section{Conclusion}
In conclusion, the much-needed rigorous benchmark L-Eval introduced in this work provides a comprehensive suite of tasks and evaluation metrics to assess the capabilities of long context language models.
We tested most of open-source LCLMs and experiments demonstrate promising gains from extending context length and gaps compared to commercial models. Our analysis using L-Eval offers valuable insights into the current state and limitations of LCLMs. We believe that with its focus on practical, long-form documents across domains, L-Eval can serve as a challenging testbed to drive advances in modeling longer contexts.

\bibliography{iclr2024_conference}

\begin{thebibliography}{54}
\providecommand{\natexlab}[1]{#1}
\providecommand{\url}[1]{\texttt{#1}}
\expandafter\ifx\csname urlstyle\endcsname\relax
  \providecommand{\doi}[1]{doi: #1}\else
  \providecommand{\doi}{doi: \begingroup \urlstyle{rm}\Url}\fi

\bibitem[Angelidis et~al.(2021)Angelidis, Amplayo, Suhara, Wang, and Lapata]{angelidis-etal-2021-extractive}
Stefanos Angelidis, Reinald~Kim Amplayo, Yoshihiko Suhara, Xiaolan Wang, and Mirella Lapata.
\newblock Extractive opinion summarization in quantized transformer spaces.
\newblock \emph{Transactions of the Association for Computational Linguistics}, 9:\penalty0 277--293, 2021.
\newblock \doi{10.1162/tacl_a_00366}.
\newblock URL \url{https://aclanthology.org/2021.tacl-1.17}.

\bibitem[Bai et~al.(2023)Bai, Lv, Zhang, Lyu, Tang, Huang, Du, Liu, Zeng, Hou, Dong, Tang, and Li]{bai2023longbench}
Yushi Bai, Xin Lv, Jiajie Zhang, Hongchang Lyu, Jiankai Tang, Zhidian Huang, Zhengxiao Du, Xiao Liu, Aohan Zeng, Lei Hou, Yuxiao Dong, Jie Tang, and Juanzi Li.
\newblock Longbench: A bilingual, multitask benchmark for long context understanding, 2023.

\bibitem[Bulatov et~al.(2023)Bulatov, Kuratov, and Burtsev]{bulatov2023scaling}
Aydar Bulatov, Yuri Kuratov, and Mikhail~S. Burtsev.
\newblock Scaling transformer to 1m tokens and beyond with rmt, 2023.

\bibitem[Chen et~al.(2022)Chen, Chu, Wiseman, and Gimpel]{chen2022summscreen}
Mingda Chen, Zewei Chu, Sam Wiseman, and Kevin Gimpel.
\newblock Summscreen: A dataset for abstractive screenplay summarization, 2022.

\bibitem[Chen et~al.(2023)Chen, Wong, Chen, and Tian]{chen2023extending}
Shouyuan Chen, Sherman Wong, Liangjian Chen, and Yuandong Tian.
\newblock Extending context window of large language models via positional interpolation, 2023.

\bibitem[Chiang et~al.(2023)Chiang, Li, Lin, Sheng, Wu, Zhang, Zheng, Zhuang, Zhuang, Gonzalez, Stoica, and Xing]{vicuna2023}
Wei-Lin Chiang, Zhuohan Li, Zi~Lin, Ying Sheng, Zhanghao Wu, Hao Zhang, Lianmin Zheng, Siyuan Zhuang, Yonghao Zhuang, Joseph~E. Gonzalez, Ion Stoica, and Eric~P. Xing.
\newblock Vicuna: An open-source chatbot impressing gpt-4 with 90\%* chatgpt quality, March 2023.
\newblock URL \url{https://lmsys.org/blog/2023-03-30-vicuna/}.

\bibitem[Chung et~al.(2018)Chung, Lee, and Glass]{chung2018supervised}
Yu-An Chung, Hung-Yi Lee, and James Glass.
\newblock Supervised and unsupervised transfer learning for question answering.
\newblock In \emph{NAACL HLT}, 2018.

\bibitem[Cobbe et~al.(2021)Cobbe, Kosaraju, Bavarian, Chen, Jun, Kaiser, Plappert, Tworek, Hilton, Nakano, Hesse, and Schulman]{cobbe2021training}
Karl Cobbe, Vineet Kosaraju, Mohammad Bavarian, Mark Chen, Heewoo Jun, Lukasz Kaiser, Matthias Plappert, Jerry Tworek, Jacob Hilton, Reiichiro Nakano, Christopher Hesse, and John Schulman.
\newblock Training verifiers to solve math word problems, 2021.

\bibitem[Dai et~al.(2019)Dai, Yang, Yang, Carbonell, Le, and Salakhutdinov]{dai-etal-2019-transformer}
Zihang Dai, Zhilin Yang, Yiming Yang, Jaime Carbonell, Quoc Le, and Ruslan Salakhutdinov.
\newblock Transformer-{XL}: Attentive language models beyond a fixed-length context.
\newblock In \emph{Proceedings of the 57th Annual Meeting of the Association for Computational Linguistics}, pp.\  2978--2988, Florence, Italy, July 2019. Association for Computational Linguistics.
\newblock \doi{10.18653/v1/P19-1285}.
\newblock URL \url{https://aclanthology.org/P19-1285}.

\bibitem[Dao et~al.(2022)Dao, Fu, Ermon, Rudra, and R\'{e}]{NEURIPS2022_67d57c32}
Tri Dao, Dan Fu, Stefano Ermon, Atri Rudra, and Christopher R\'{e}.
\newblock Flashattention: Fast and memory-efficient exact attention with io-awareness.
\newblock In S.~Koyejo, S.~Mohamed, A.~Agarwal, D.~Belgrave, K.~Cho, and A.~Oh (eds.), \emph{Advances in Neural Information Processing Systems}, volume~35, pp.\  16344--16359. Curran Associates, Inc., 2022.
\newblock URL \url{https://proceedings.neurips.cc/paper_files/paper/2022/file/67d57c32e20fd0a7a302cb81d36e40d5-Paper-Conference.pdf}.

\bibitem[Dasigi et~al.(2021)Dasigi, Lo, Beltagy, Cohan, Smith, and Gardner]{dasigi2021dataset}
Pradeep Dasigi, Kyle Lo, Iz~Beltagy, Arman Cohan, Noah~A. Smith, and Matt Gardner.
\newblock A dataset of information-seeking questions and answers anchored in research papers, 2021.

\bibitem[Ding et~al.(2023)Ding, Ma, Dong, Zhang, Huang, Wang, Zheng, and Wei]{ding2023longnet}
Jiayu Ding, Shuming Ma, Li~Dong, Xingxing Zhang, Shaohan Huang, Wenhui Wang, Nanning Zheng, and Furu Wei.
\newblock Longnet: Scaling transformers to 1,000,000,000 tokens, 2023.

\bibitem[Du et~al.(2022)Du, Qian, Liu, Ding, Qiu, Yang, and Tang]{du2022glm}
Zhengxiao Du, Yujie Qian, Xiao Liu, Ming Ding, Jiezhong Qiu, Zhilin Yang, and Jie Tang.
\newblock Glm: General language model pretraining with autoregressive blank infilling.
\newblock In \emph{Proceedings of the 60th Annual Meeting of the Association for Computational Linguistics (Volume 1: Long Papers)}, pp.\  320--335, 2022.

\bibitem[Dubois et~al.(2023)Dubois, Li, Taori, Zhang, Gulrajani, Ba, Guestrin, Liang, and Hashimoto]{dubois2023alpacafarm}
Yann Dubois, Xuechen Li, Rohan Taori, Tianyi Zhang, Ishaan Gulrajani, Jimmy Ba, Carlos Guestrin, Percy Liang, and Tatsunori~B. Hashimoto.
\newblock Alpacafarm: A simulation framework for methods that learn from human feedback, 2023.

\bibitem[Fabbri et~al.(2019)Fabbri, Li, She, Li, and Radev]{fabbri2019multinews}
Alexander~R. Fabbri, Irene Li, Tianwei She, Suyi Li, and Dragomir~R. Radev.
\newblock Multi-news: a large-scale multi-document summarization dataset and abstractive hierarchical model, 2019.

\bibitem[Feng et~al.(2021)Feng, Patel, Wan, and Joshi]{Feng_2021}
Song Feng, Siva~Sankalp Patel, Hui Wan, and Sachindra Joshi.
\newblock {MultiDoc}2dial: Modeling dialogues grounded in multiple documents.
\newblock In \emph{Proceedings of the 2021 Conference on Empirical Methods in Natural Language Processing}. Association for Computational Linguistics, 2021.
\newblock \doi{10.18653/v1/2021.emnlp-main.498}.
\newblock URL \url{https://doi.org/10.18653%2Fv1%2F2021.emnlp-main.498}.

\bibitem[Fu et~al.(2023)Fu, Dao, Saab, Thomas, Rudra, and Re]{fu2023hungry}
Daniel~Y Fu, Tri Dao, Khaled~Kamal Saab, Armin~W Thomas, Atri Rudra, and Christopher Re.
\newblock Hungry hungry hippos: Towards language modeling with state space models.
\newblock In \emph{The Eleventh International Conference on Learning Representations}, 2023.
\newblock URL \url{https://openreview.net/forum?id=COZDy0WYGg}.

\bibitem[Hendrycks et~al.(2021{\natexlab{a}})Hendrycks, Burns, Basart, Zou, Mazeika, Song, and Steinhardt]{hendrycks2021measuring}
Dan Hendrycks, Collin Burns, Steven Basart, Andy Zou, Mantas Mazeika, Dawn Song, and Jacob Steinhardt.
\newblock Measuring massive multitask language understanding, 2021{\natexlab{a}}.

\bibitem[Hendrycks et~al.(2021{\natexlab{b}})Hendrycks, Burns, Chen, and Ball]{hendrycks2021cuad}
Dan Hendrycks, Collin Burns, Anya Chen, and Spencer Ball.
\newblock Cuad: An expert-annotated nlp dataset for legal contract review, 2021{\natexlab{b}}.

\bibitem[Huang et~al.(2021)Huang, Cao, Parulian, Ji, and Wang]{huang-etal-2021-efficient}
Luyang Huang, Shuyang Cao, Nikolaus Parulian, Heng Ji, and Lu~Wang.
\newblock Efficient attentions for long document summarization.
\newblock In \emph{Proceedings of the 2021 Conference of the North American Chapter of the Association for Computational Linguistics: Human Language Technologies}, pp.\  1419--1436, Online, June 2021. Association for Computational Linguistics.
\newblock \doi{10.18653/v1/2021.naacl-main.112}.
\newblock URL \url{https://aclanthology.org/2021.naacl-main.112}.

\bibitem[Kočiský et~al.(2017)Kočiský, Schwarz, Blunsom, Dyer, Hermann, Melis, and Grefenstette]{kočiský2017narrativeqa}
Tomáš Kočiský, Jonathan Schwarz, Phil Blunsom, Chris Dyer, Karl~Moritz Hermann, Gábor Melis, and Edward Grefenstette.
\newblock The narrativeqa reading comprehension challenge, 2017.

\bibitem[Kwiatkowski et~al.(2019)Kwiatkowski, Palomaki, Redfield, Collins, Parikh, Alberti, Epstein, Polosukhin, Devlin, Lee, Toutanova, Jones, Kelcey, Chang, Dai, Uszkoreit, Le, and Petrov]{kwiatkowski-etal-2019-natural}
Tom Kwiatkowski, Jennimaria Palomaki, Olivia Redfield, Michael Collins, Ankur Parikh, Chris Alberti, Danielle Epstein, Illia Polosukhin, Jacob Devlin, Kenton Lee, Kristina Toutanova, Llion Jones, Matthew Kelcey, Ming-Wei Chang, Andrew~M. Dai, Jakob Uszkoreit, Quoc Le, and Slav Petrov.
\newblock Natural questions: A benchmark for question answering research.
\newblock \emph{Transactions of the Association for Computational Linguistics}, 7:\penalty0 452--466, 2019.
\newblock \doi{10.1162/tacl_a_00276}.
\newblock URL \url{https://aclanthology.org/Q19-1026}.

\bibitem[Li et~al.(2023{\natexlab{a}})Li, Shao, Xie, Sheng, Zheng, Gonzalez, Stoica, Ma, and Zhang]{longchat2023}
Dacheng Li, Rulin Shao, Anze Xie, Ying Sheng, Lianmin Zheng, Joseph~E. Gonzalez, Ion Stoica, Xuezhe Ma, and Hao Zhang.
\newblock How long can open-source llms truly promise on context length?, June 2023{\natexlab{a}}.
\newblock URL \url{https://lmsys.org/blog/2023-06-29-longchat}.

\bibitem[Li et~al.(2023{\natexlab{b}})Li, Gong, Feng, Xu, Zhang, Wu, and Kong]{li2023incontext}
Mukai Li, Shansan Gong, Jiangtao Feng, Yiheng Xu, Jun Zhang, Zhiyong Wu, and Lingpeng Kong.
\newblock In-context learning with many demonstration examples, 2023{\natexlab{b}}.

\bibitem[Li et~al.(2023{\natexlab{c}})Li, Zhang, Dubois, Taori, Gulrajani, Guestrin, Liang, and Hashimoto]{alpaca_eval}
Xuechen Li, Tianyi Zhang, Yann Dubois, Rohan Taori, Ishaan Gulrajani, Carlos Guestrin, Percy Liang, and Tatsunori~B. Hashimoto.
\newblock Alpacaeval: An automatic evaluator of instruction-following models.
\newblock \url{https://github.com/tatsu-lab/alpaca_eval}, 2023{\natexlab{c}}.

\bibitem[Liang et~al.(2023)Liang, Wang, Huang, Wu, Wu, Lu, Ma, and Li]{liang2023unleashing}
Xinnian Liang, Bing Wang, Hui Huang, Shuangzhi Wu, Peihao Wu, Lu~Lu, Zejun Ma, and Zhoujun Li.
\newblock Unleashing infinite-length input capacity for large-scale language models with self-controlled memory system.
\newblock \emph{arXiv preprint arXiv:2304.13343}, 2023.

\bibitem[Lin(2004)]{lin-2004-rouge}
Chin-Yew Lin.
\newblock {ROUGE}: A package for automatic evaluation of summaries.
\newblock In \emph{Text Summarization Branches Out}, pp.\  74--81, Barcelona, Spain, July 2004. Association for Computational Linguistics.
\newblock URL \url{https://aclanthology.org/W04-1013}.

\bibitem[Liu et~al.(2023)Liu, Lin, Hewitt, Paranjape, Bevilacqua, Petroni, and Liang]{liu2023lost}
Nelson~F. Liu, Kevin Lin, John Hewitt, Ashwin Paranjape, Michele Bevilacqua, Fabio Petroni, and Percy Liang.
\newblock Lost in the middle: How language models use long contexts, 2023.

\bibitem[LocalLLaMA(2023{\natexlab{a}})]{dynamicNTK}
LocalLLaMA.
\newblock Dynamically scaled rope further increases performance of long context llama with zero fine-tuning, July 2023{\natexlab{a}}.
\newblock URL \url{https://www.reddit.com/r/LocalLLaMA/comments/14mrgpr/dynamically_scaled_rope_further_increases/}.

\bibitem[LocalLLaMA(2023{\natexlab{b}})]{fixedNTK}
LocalLLaMA.
\newblock Ntk-aware scaled rope allows llama models to have extended (8k+) context size without any fine-tuning and minimal perplexity degradation., June 2023{\natexlab{b}}.
\newblock URL \url{https://www.reddit.com/r/LocalLLaMA/comments/14lz7j5/ntkaware_scaled_rope_allows_llama_models_to_have/}.

\bibitem[Mohtashami \& Jaggi(2023)Mohtashami and Jaggi]{mohtashami2023landmark}
Amirkeivan Mohtashami and Martin Jaggi.
\newblock Landmark attention: Random-access infinite context length for transformers.
\newblock \emph{arXiv preprint arXiv:2305.16300}, 2023.

\bibitem[Nijkamp et~al.(2023)Nijkamp, Xie, Hayashi, Pang, Xia, Xing, Vig, Yavuz, Laban, Krause, Purushwalkam, Niu, Kryscinski, Murakhovs'ka, Choubey, Fabbri, Liu, Meng, Tu, Bhat, Wu, Savarese, Zhou, Joty, and Xiong]{XGen}
Erik Nijkamp, Tian Xie, Hiroaki Hayashi, Bo~Pang, Congying Xia, Chen Xing, Jesse Vig, Semih Yavuz, Philippe Laban, Ben Krause, Senthil Purushwalkam, Tong Niu, Wojciech Kryscinski, Lidiya Murakhovs'ka, Prafulla~Kumar Choubey, Alex Fabbri, Ye~Liu, Rui Meng, Lifu Tu, Meghana Bhat, Chien-Sheng Wu, Silvio Savarese, Yingbo Zhou, Shafiq~Rayhan Joty, and Caiming Xiong.
\newblock Long sequence modeling with xgen: A 7b llm trained on 8k input sequence length.
\newblock Salesforce AI Research Blog, 2023.
\newblock URL \url{https://blog.salesforceairesearch.com/xgen}.

\bibitem[Pang et~al.(2022)Pang, Parrish, Joshi, Nangia, Phang, Chen, Padmakumar, Ma, Thompson, He, and Bowman]{pang2022quality}
Richard~Yuanzhe Pang, Alicia Parrish, Nitish Joshi, Nikita Nangia, Jason Phang, Angelica Chen, Vishakh Padmakumar, Johnny Ma, Jana Thompson, He~He, and Samuel~R. Bowman.
\newblock Quality: Question answering with long input texts, yes!, 2022.

\bibitem[Peng et~al.(2023{\natexlab{a}})Peng, Alcaide, Anthony, Albalak, Arcadinho, Cao, Cheng, Chung, Grella, GV, He, Hou, Kazienko, Kocon, Kong, Koptyra, Lau, Mantri, Mom, Saito, Tang, Wang, Wind, Wozniak, Zhang, Zhang, Zhao, Zhou, Zhu, and Zhu]{peng2023rwkv}
Bo~Peng, Eric Alcaide, Quentin Anthony, Alon Albalak, Samuel Arcadinho, Huanqi Cao, Xin Cheng, Michael Chung, Matteo Grella, Kranthi~Kiran GV, Xuzheng He, Haowen Hou, Przemyslaw Kazienko, Jan Kocon, Jiaming Kong, Bartlomiej Koptyra, Hayden Lau, Krishna Sri~Ipsit Mantri, Ferdinand Mom, Atsushi Saito, Xiangru Tang, Bolun Wang, Johan~S. Wind, Stansilaw Wozniak, Ruichong Zhang, Zhenyuan Zhang, Qihang Zhao, Peng Zhou, Jian Zhu, and Rui-Jie Zhu.
\newblock Rwkv: Reinventing rnns for the transformer era, 2023{\natexlab{a}}.

\bibitem[Peng et~al.(2023{\natexlab{b}})Peng, Quesnelle, Fan, and Shippole]{peng2023yarn}
Bowen Peng, Jeffrey Quesnelle, Honglu Fan, and Enrico Shippole.
\newblock Yarn: Efficient context window extension of large language models, 2023{\natexlab{b}}.

\bibitem[Press et~al.(2022)Press, Smith, and Lewis]{press2022train}
Ofir Press, Noah Smith, and Mike Lewis.
\newblock Train short, test long: Attention with linear biases enables input length extrapolation.
\newblock In \emph{International Conference on Learning Representations}, 2022.
\newblock URL \url{https://openreview.net/forum?id=R8sQPpGCv0}.

\bibitem[Qin et~al.(2023)Qin, Sun, Lu, Deng, Li, Han, Dai, Kong, and Zhong]{qin2023linear}
Zhen Qin, Weixuan Sun, Kaiyue Lu, Hui Deng, Dongxu Li, Xiaodong Han, Yuchao Dai, Lingpeng Kong, and Yiran Zhong.
\newblock Linearized relative positional encoding.
\newblock \emph{CoRR}, abs/2307.09270, 2023.
\newblock \doi{10.48550/arXiv.2307.09270}.
\newblock URL \url{https://doi.org/10.48550/arXiv.2307.09270}.

\bibitem[Shaham et~al.(2022)Shaham, Segal, Ivgi, Efrat, Yoran, Haviv, Gupta, Xiong, Geva, Berant, and Levy]{shaham2022scrolls}
Uri Shaham, Elad Segal, Maor Ivgi, Avia Efrat, Ori Yoran, Adi Haviv, Ankit Gupta, Wenhan Xiong, Mor Geva, Jonathan Berant, and Omer Levy.
\newblock Scrolls: Standardized comparison over long language sequences, 2022.

\bibitem[Shaham et~al.(2023)Shaham, Ivgi, Efrat, Berant, and Levy]{shaham2023zeroscrolls}
Uri Shaham, Maor Ivgi, Avia Efrat, Jonathan Berant, and Omer Levy.
\newblock Zeroscrolls: A zero-shot benchmark for long text understanding, 2023.

\bibitem[Sharma et~al.(2019)Sharma, Li, and Wang]{sharma-etal-2019-bigpatent}
Eva Sharma, Chen Li, and Lu~Wang.
\newblock {BIGPATENT}: A large-scale dataset for abstractive and coherent summarization.
\newblock In \emph{Proceedings of the 57th Annual Meeting of the Association for Computational Linguistics}, pp.\  2204--2213, Florence, Italy, July 2019. Association for Computational Linguistics.
\newblock \doi{10.18653/v1/P19-1212}.
\newblock URL \url{https://aclanthology.org/P19-1212}.

\bibitem[Su et~al.(2022)Su, Lu, Pan, Murtadha, Wen, and Liu]{su2022roformer}
Jianlin Su, Yu~Lu, Shengfeng Pan, Ahmed Murtadha, Bo~Wen, and Yunfeng Liu.
\newblock Roformer: Enhanced transformer with rotary position embedding, 2022.

\bibitem[Sun et~al.(2021)Sun, Krishna, Mattarella-Micke, and Iyyer]{sun2021long}
Simeng Sun, Kalpesh Krishna, Andrew Mattarella-Micke, and Mohit Iyyer.
\newblock Do long-range language models actually use long-range context?
\newblock In \emph{Proceedings of the 2021 Conference on Empirical Methods in Natural Language Processing}, pp.\  807--822, 2021.

\bibitem[Sun et~al.(2022)Sun, Dong, Patra, Ma, Huang, Benhaim, Chaudhary, Song, and Wei]{sun2022lengthextrapolatable}
Yutao Sun, Li~Dong, Barun Patra, Shuming Ma, Shaohan Huang, Alon Benhaim, Vishrav Chaudhary, Xia Song, and Furu Wei.
\newblock A length-extrapolatable transformer, 2022.

\bibitem[Sun et~al.(2023)Sun, Dong, Huang, Ma, Xia, Xue, Wang, and Wei]{sun2023retentive}
Yutao Sun, Li~Dong, Shaohan Huang, Shuming Ma, Yuqing Xia, Jilong Xue, Jianyong Wang, and Furu Wei.
\newblock Retentive network: A successor to transformer for large language models, 2023.

\bibitem[Suzgun et~al.(2022)Suzgun, Scales, Schärli, Gehrmann, Tay, Chung, Chowdhery, Le, Chi, Zhou, and Wei]{suzgun2022challenging}
Mirac Suzgun, Nathan Scales, Nathanael Schärli, Sebastian Gehrmann, Yi~Tay, Hyung~Won Chung, Aakanksha Chowdhery, Quoc~V. Le, Ed~H. Chi, Denny Zhou, and Jason Wei.
\newblock Challenging big-bench tasks and whether chain-of-thought can solve them, 2022.

\bibitem[Tay et~al.(2020)Tay, Dehghani, Abnar, Shen, Bahri, Pham, Rao, Yang, Ruder, and Metzler]{tay2020long}
Yi~Tay, Mostafa Dehghani, Samira Abnar, Yikang Shen, Dara Bahri, Philip Pham, Jinfeng Rao, Liu Yang, Sebastian Ruder, and Donald Metzler.
\newblock Long range arena: A benchmark for efficient transformers, 2020.

\bibitem[Touvron et~al.(2023{\natexlab{a}})Touvron, Lavril, Izacard, Martinet, Lachaux, Lacroix, Rozière, Goyal, Hambro, Azhar, Rodriguez, Joulin, Grave, and Lample]{touvron2023llama}
Hugo Touvron, Thibaut Lavril, Gautier Izacard, Xavier Martinet, Marie-Anne Lachaux, Timothée Lacroix, Baptiste Rozière, Naman Goyal, Eric Hambro, Faisal Azhar, Aurelien Rodriguez, Armand Joulin, Edouard Grave, and Guillaume Lample.
\newblock Llama: Open and efficient foundation language models, 2023{\natexlab{a}}.

\bibitem[Touvron et~al.(2023{\natexlab{b}})Touvron, Martin, Stone, Albert, Almahairi, Babaei, Bashlykov, Batra, Bhargava, Bhosale, Bikel, Blecher, Ferrer, Chen, Cucurull, Esiobu, Fernandes, Fu, Fu, Fuller, Gao, Goswami, Goyal, Hartshorn, Hosseini, Hou, Inan, Kardas, Kerkez, Khabsa, Kloumann, Korenev, Koura, Lachaux, Lavril, Lee, Liskovich, Lu, Mao, Martinet, Mihaylov, Mishra, Molybog, Nie, Poulton, Reizenstein, Rungta, Saladi, Schelten, Silva, Smith, Subramanian, Tan, Tang, Taylor, Williams, Kuan, Xu, Yan, Zarov, Zhang, Fan, Kambadur, Narang, Rodriguez, Stojnic, Edunov, and Scialom]{touvron2023llama2}
Hugo Touvron, Louis Martin, Kevin Stone, Peter Albert, Amjad Almahairi, Yasmine Babaei, Nikolay Bashlykov, Soumya Batra, Prajjwal Bhargava, Shruti Bhosale, Dan Bikel, Lukas Blecher, Cristian~Canton Ferrer, Moya Chen, Guillem Cucurull, David Esiobu, Jude Fernandes, Jeremy Fu, Wenyin Fu, Brian Fuller, Cynthia Gao, Vedanuj Goswami, Naman Goyal, Anthony Hartshorn, Saghar Hosseini, Rui Hou, Hakan Inan, Marcin Kardas, Viktor Kerkez, Madian Khabsa, Isabel Kloumann, Artem Korenev, Punit~Singh Koura, Marie-Anne Lachaux, Thibaut Lavril, Jenya Lee, Diana Liskovich, Yinghai Lu, Yuning Mao, Xavier Martinet, Todor Mihaylov, Pushkar Mishra, Igor Molybog, Yixin Nie, Andrew Poulton, Jeremy Reizenstein, Rashi Rungta, Kalyan Saladi, Alan Schelten, Ruan Silva, Eric~Michael Smith, Ranjan Subramanian, Xiaoqing~Ellen Tan, Binh Tang, Ross Taylor, Adina Williams, Jian~Xiang Kuan, Puxin Xu, Zheng Yan, Iliyan Zarov, Yuchen Zhang, Angela Fan, Melanie Kambadur, Sharan Narang, Aurelien Rodriguez, Robert Stojnic, Sergey Edunov, and Thomas
  Scialom.
\newblock Llama 2: Open foundation and fine-tuned chat models, 2023{\natexlab{b}}.

\bibitem[Tseng et~al.(2016)Tseng, Shen, Lee, and Lee]{tseng2016towards}
Bo-Hsiang Tseng, Sheng-Syun Shen, Hung-Yi Lee, and Lin-Shan Lee.
\newblock Towards machine comprehension of spoken content: Initial toefl listening comprehension test by machine.
\newblock In \emph{INTERSPEECH}, 2016.

\bibitem[Wang et~al.(2023)Wang, Zhang, Xie, Yao, Tian, Wang, Xi, Cheng, Liu, Zheng, and Chen]{wang2023easyedit}
Peng Wang, Ningyu Zhang, Xin Xie, Yunzhi Yao, Bozhong Tian, Mengru Wang, Zekun Xi, Siyuan Cheng, Kangwei Liu, Guozhou Zheng, and Huajun Chen.
\newblock Easyedit: An easy-to-use knowledge editing framework for large language models, 2023.

\bibitem[Yuan et~al.(2021)Yuan, Liu, and Neubig]{yuan2021automate}
Weizhe Yuan, Pengfei Liu, and Graham Neubig.
\newblock Can we automate scientific reviewing?, 2021.

\bibitem[Zhang et~al.(2023)Zhang, Jiang, Feng, Zheng, and Kong]{zhang2023cab}
Jun Zhang, Shuyang Jiang, Jiangtao Feng, Lin Zheng, and Lingpeng Kong.
\newblock Cab: Comprehensive attention benchmarking on long sequence modeling, 2023.

\bibitem[Zheng et~al.(2023)Zheng, Chiang, Sheng, Zhuang, Wu, Zhuang, Lin, Li, Li, Xing, Zhang, Gonzalez, and Stoica]{zheng2023judging}
Lianmin Zheng, Wei-Lin Chiang, Ying Sheng, Siyuan Zhuang, Zhanghao Wu, Yonghao Zhuang, Zi~Lin, Zhuohan Li, Dacheng Li, Eric.~P Xing, Hao Zhang, Joseph~E. Gonzalez, and Ion Stoica.
\newblock Judging llm-as-a-judge with mt-bench and chatbot arena, 2023.

\bibitem[Zhong et~al.(2021)Zhong, Yin, Yu, Zaidi, Mutuma, Jha, Awadallah, Celikyilmaz, Liu, Qiu, and Radev]{zhong2021qmsum}
Ming Zhong, Da~Yin, Tao Yu, Ahmad Zaidi, Mutethia Mutuma, Rahul Jha, Ahmed~Hassan Awadallah, Asli Celikyilmaz, Yang Liu, Xipeng Qiu, and Dragomir Radev.
\newblock Qmsum: A new benchmark for query-based multi-domain meeting summarization, 2021.

\end{thebibliography}
\bibliographystyle{iclr2024_conference}

\newpage
\appendix
\section{Appendix}
\subsection{Baseline Models in L-Eval}\label{sec:baselines_appendix}
\paragraph{Commercial Models}
\begin{itemize}
    \item {Claude-100k} is developed by Anthropic\footnote{\url{https://www.anthropic.com/index/100k-context-windows}} and targets understanding extremely long documents and answering related questions. It has the longest context length among all the LLMs.
    \item GPT-4-32k is developed by OpenAI\footnote{\url{https://platform.openai.com/docs/models/gpt-4}}. It is the long context version of GPT-4 maintaining very strong reasoning ability over 32k context length but also the most expensive model.
    \item {Turbo-4k-0613} is the snapshot of {GPT-3.5}\footnote{\url{https://platform.openai.com/docs/models/gpt-3-5}} from June 13th 2023 which can handle up to 4k input tokens. {Turbo-16k-0613} is the released long context version of Turbo-4k-0613.
\end{itemize}

\paragraph{Open-source Models}
\begin{itemize}
    \item {Llama1}~\citep{touvron2023llama}\footnote{\url{https://github.com/facebookresearch/llama}} is a widely used open-source model developed by Meta AI with a 2k pre-training context length. The first version of Llama did not release a chatbot-based model.
    \item {Vicuna1.3}~\citep{vicuna2023}~\footnote{\url{https://github.com/lm-sys/FastChat.git}} is a chatbot fine-tuned from Llama1 on shareGPT.
    \item Longchat-16k~\citep{longchat2023}~\footnote{\url{https://github.com/DachengLi1/LongChat}} is the long context version of Vicuna. It uses positional interpolation to adapt 16k context. Concretely, they further fine-tune Llama1 on lengthy dialogues (16k tokens) from shareGPT.
    \item Llama2~\citep{touvron2023llama2} is the second version of Llama recently released by Meta AI. The updated version has 4k pretraining context with more powerful long context understanding capabilities.
    \item Llama2-chat~\citep{touvron2023llama2} is a chatbot based on Llama2 released together with Llama2. Please notice that if we do not follow the pre-defined input format, i.e., ignore the special tokens, there will be a significant degradation in performance.
    \item Llama2-NTK-chat~\citep{fixedNTK} is the long context version of Llama2-chat. It uses NTK-aware positional embedding. If we want to extend the model context window to $t$ (we call $t$ as a scale-up factor) times its original pertaining size, we just need to increase the original base=10,000 of RoPE ($\theta_n={10000}^{-2n/d}$) to 10,000 $\times t^{\frac{d}{d-2}}$ where $d$ is the head dimension in Transformer. In our experiments, this theory does not hold in practical tasks (see section \S\ref{sec:analysis_app}), which means the model still tends to generate random tokens when setting $t=4$ on 16k context length.  We set $t=8$ in experiments.
    \item Llama2-NTK-chat (Dyn)~\citep{dynamicNTK} is the dynamic version of Llama2-NTK-chat. The only difference is that the scale-up factor $t$ in dynamic NTK depends on the current input length $L$ and the pertaining length $l$, i.e., $t=\frac{L}{l}$ 
    \item Vicuna1.5-16k uses Llama2 as the base model and performs further finetuning on concatenated 16k tokens lengthy dialogues from shareGPT. This model is based on positional interpolation which helps the training process converge fast.
    \item LongChat1.5-32k is the 32k version of Vicuna1.5-16k.
    \item Chatglm2-8k~\citep{du2022glm}\footnote{\url{https://github.com/THUDM/ChatGLM2-6B}}  is the second version of the open-source bilingual chat model Chatglm. The context length of the base model is further pretrained with 32k context window and finetuned on dialogue data with 8k context window. 
    \item Chatglm2-32k is the long context version of Chatglm2 using positional interpolation.
    \item XGen-8k-inst\footnote{\url{https://github.com/salesforce/xgen}} developed by salesforce follows a multi-stage pertaining procedure. They first train the model with 2k context length and progressively increase the pertaining length to 4k, finally reaching 8k.
    \item {MPT-7B-StoryWriter-65k}\footnote{\url{https://www.mosaicml.com/blog/mpt-7b}} is designed to handle super-long context lengths. It was tuned on MPT-7B with a context length of 65k tokens on a subset of Books3 dataset.
\end{itemize}

\subsection{Human Evaluation}
\label{sec:appendix:human-eval}
\begin{table}
\centering
\resizebox{1.0\textwidth}{!}{
\begin{tabular}{lccccc|ccccc}
\toprule
\textbf{Model} & \textbf{\#Level-1} & \textbf{\#Level-2} & \textbf{\#Level-3} &  \textbf{\#Level-4} &  \textbf{\#Level-5}  &\bf Human-Avg & \bf GPT-4 & \bf GPT-3.5 & \bf F-1  & \bf R-L \\
\midrule
\rowcolor{mypink!30}
\multicolumn{11}{c}{\textit{Length-instruction-enhanced evaluation reulsts}} \\
\midrule
llama2-7b-chat & 53 & 38 & 74 & 46 & 44 & 2.96 & 38.52 & 42.37 & 24.26 & 28.48 \\
llama2-13b-chat & 41 & 37 & 68 & 59 & 50 & 3.15 & 40.00 & 48.07 & 26.10 & 30.90 \\
Turbo-4k-0613 &43 & 29 & 51 & 72 & 60 & 3.30 & 42.05 & 43.75 & 26.05 & 30.75 \\
\midrule
Claude-100k &14 & 15 & 37 & 69 & \bf 120 & \bf 4.04 & \bf 60.88 & \bf 63.75 & 26.39 & 31.57 \\
turbo-16k-0613 &37 & 12 & 43 & \bf 90 & 73 & 3.58 & 50.00 & 50.00 & \bf 27.99 & \bf 32.93 \\
Vicuan-7b-16k & \color{red}{125} & 26 & 45 & 43 & 16 & \color{red}{2.21} & \color{red}{23.23} & 35.09 & 16.25 & 19.40 \\
longchat-7b-16k  & \color{red}{113} & 29 & 61 & 32 & 20 & \color{red}{2.28} & \color{red}{23.82} & 37.57 & 17.12 & 20.81\\
\midrule
\rowcolor{mypink!30}
\multicolumn{11}{c}{\textit{Original evaluation results}} \\
\midrule
llama2-7b-chat & \color{red}{136} & 49 & 47 & 15 & 8 & 1.86 & 32.35 & 42.40 & 14.29 & 17.72 \\
llama2-13b-chat & 92 & 50 & 64 & 38 & 11 & 2.31 & 35.00 & 55.76 & 13.62 & 18.10 \\
Turbo-4k-0613 & 66 & 38 & 60 & 40 & 51 & 2.89 & 50.00 & 44.06 & 20.06 & 24.88 \\
\midrule
Claude-100k & 27 & 52 & 81 & 66 & 29 & \textbf{3.08} & \bf 53.23 & \bf 76.68 & 15.31 & 19.59 \\
turbo-16k-0613 & 42 & 40 & 78 & 64 & 31 & 3.00 & 50.00 & 50.00 & 20.60 & \bf 25.96 \\
Vicuan-7b-16k & \color{red}{138} & 49 & 46 & 14 & 8 & \color{red}{1.84} &\color{red}{23.23} & 38.27 & 14.69 & 17.90 \\
longchat-7b-16k & \color{red}{156} & 40 & 36 & 18 & 5 & \color{red}{1.72} & \color{red}{22.05} & 35.76 & 13.25 & 15.73 \\

\bottomrule
\end{tabular}
}
\caption{Human evaluation results and results from other automatic metrics where \textbf{\#Level-N} denotes the number of outputs (the sum from all annotators) in Level-N on the 85-question subset. Texts colored with red mean very unsatisfactory results. }
\label{tab:human_eval}
\end{table} 

Evaluating long-sequence, open-ended tasks remains a challenge. As previously discussed, almost all metrics, including the highly accurate automatic metric, the GPT-4 evaluator, exhibit bias. Consequently, human evaluation may be the most equitable metric to assess these models. In this section, we detail the human evaluation procedure conducted on seven baseline models using an 85-question subset. Our goal is to examine the correlation between human judgement and automatic metrics. Additionally, we aim to evaluate the performance of the length-instruction-enhanced evaluation method proposed in this paper.

\paragraph{Experimental setup}
% We evaluate 7 models including 3 commercial models and 4 open-source models: (1) Claude-100k, (2) turbo-16k-0613, (3) Turbo-4k-0613, (4) Vicuna1.5-7b-16k (Llama2) (5) Longchat-7b-16k (Llama1) (6) Llama2-7b-chat (7) Llama2-13b-chat on the 85-question subset from L-Eval open-ended tasks. Each sample is scored by three annotators (Ph.D. students researching on long context language models). We then calculate the average score as the final result. We use the Kendall-Tau correlation coefficient to measure whether the ranking given by these automatic metrics can correlate with the ranking given by the annotators. We let the models generate twice where the model first generates outputs in the origin mode, i.e., no length instruction given, and then the models are decoding with the given length instructions. We use greedy search as the decoding algorithm to reduce variance. We rank the model outputs in five levels:
We evaluate seven models, comprising three commercial and four open-source models: (1) Claude-100k, (2) turbo-16k-0613, (3) Turbo-4k-0613, (4) Vicuna1.5-7b-16k (Llama2), (5) Longchat-7b-16k (Llama1), (6) Llama2-7b-chat, and (7) Llama2-13b-chat. These models are tested on an 85-question subset from L-Eval open-ended tasks. Each sample is scored by three annotators, all of whom are Ph.D. students researching long context language models. We calculate the average score to obtain the final human evalution results. To determine if the ranking produced by these automatic metrics correlates with the ranking provided by the annotators, we use the Kendall-Tau correlation coefficient. We allow each model to generate outputs twice: first in the original mode without any length instruction, and then with the given length instructions. To minimize variance, we use greedy search as the decoding algorithm. The model outputs are ranked on a five-level scale:
\begin{itemize}
    \item Level-1 (worst): The response is totally unhelpful to answer the question. 
    \item Level-2: The output generally deviates from the original question, but some information is useful to solve the problem.
    \item Level-3: The response is partially correct, but the generated answer may contain some errors or omit key information.
    \item Level-4: Most of the response is correct, but there may be minor issues such as being overly long (which cannot be considered a flaw if it is a reasonable explanation), or it might omit some information, but this does not affect the overall meaning.
    \item Level-5 (best): The output is close-to-human or even better.
\end{itemize}

\paragraph{Human evaluation results}
% Results of our human evaluation are shown in Table~\ref{tab:human_eval}. As we can see, although open-source long context models have been fine-tuned on longer context, they still struggle when facing very long input sequences during inference. The number of Level-1 outputs is greatly increased when feeding numerous input tokens,  while the LLMs with only 4k pertaining context length, although not achieving a high score, can still maintain their generation quality at a partially correct level. We can also observe that the N-gram metrics F-1 and ROUGE generally can not correlate with the human evaluation results. Considering it is still impractical to test numerous samples due to the high cost and inefficiency, we will also release our human assessment to facilitate research on the metrics. 
The results of our human evaluation are presented in Table~\ref{tab:human_eval}. As can be seen, despite being fine-tuned on longer contexts, open-source long context models still struggle with very long input sequences during inference. When fed with numerous input tokens, the number of Level-1 outputs from open-source LCLMs significantly increases, while the LLMs with only a 4k context length can maintain their generation quality at a partially correct level, albeit without achieving high scores. It's also observable that the N-gram metrics F-1 and ROUGE generally do not correlate with the human evaluation results. Given the impracticality of testing a large number of samples using LLMs due to high costs and inefficiency, we also urge for more advanced metrics. We will release our human assessment to aid research on these metrics.

\subsection{Analysis}\label{sec:analysis_app}
\begin{table}[h!]
\small
\setlength{\tabcolsep}{1.8mm}
\renewcommand\arraystretch{1.1}

\begin{tabular}{@{}lccccccccc@{}}
\toprule
\textbf{Model} & \textbf{Ret.} & \textbf{Tokens} & \textbf{Fin.} & \textbf{Contract} & \textbf{Multidoc} & \textbf{Nrtv} & \textbf{NQ} & \textbf{SCI} & \textbf{Avg. } \\
\midrule
Turbo-16k-0613 & \xmark & \cellcolor{gray!40}16k & \textbf{45.36} & 24.87 & 31.45 & 18.20 & 45.90 & 28.25 & 32.33 \\
AdaEmb-Turbo-0613 & \cmark & \cellcolor{gray!15} 4k & 39.69 & 24.09 & \textbf{35.62} & \textbf{18.59} & 49.66 & \textbf{33.36}  & \textbf{33.50} \\
BM25-Turbo-0613 & \cmark & \cellcolor{gray!15} 4k & 40.79 & \textbf{26.10} & 35.17 & 16.32 & \textbf{53.73} & 25.83  & 32.99 \\
\midrule
\rowcolor{mypink!30}
\multicolumn{10}{c}{\textit{Truncating input tokens to the pretraining context length}} \\
\midrule

Llama2-7b-chat & \xmark & \cellcolor{gray!15}4k & \textbf{40.06} & 23.00 & \textbf{27.28} & 13.48 & 28.11 & {25.95} &  {26.31} \\
Llama2-13b-chat & \xmark & \cellcolor{gray!15}4k & 38.07 & \bf23.14 & 26.14 & \bf16.76 & \textbf{35.43} & \textbf{27.46} &  \textbf{27.83} \\
Vicuna1.3-7b & \xmark & \cellcolor{gray!5}2k & 30.49 & 17.69 & 17.70 & 14.57 & 15.49 & 7.69 &  17.27 \\
Longchat-7b-16k & \xmark & \cellcolor{gray!5}2k & 27.27 & 19.78 & 13.99 & 13.21 & 18.11 & 7.61 &  16.66 \\
Chatglm2-6b-8k & \xmark & \cellcolor{gray!5}2k & 29.60 & 19.06 & 16.22 & 13.21 &17.52& 12.26 &  17.97 \\
XGen-7b-8k (2k-4k-8k) & \xmark & \cellcolor{gray!5}2k & {34.43} & {21.28} &  {21.59} & {14.97} & {29.58} & {14.12} &  {22.66} \\

\midrule
\rowcolor{mypink!30}
\multicolumn{10}{c}{\textit{Truncating input tokens to the further finetuning context length}} \\
\midrule
Chatglm2-7b-32k & \xmark &\cellcolor{gray!60}32k & 30.27 & \bf26.95 & \bf24.97 & 14.00 & \bf37.94 & 26.44 & \bf 26.76 \\
Longchat1.5-7b-32k & \xmark &\cellcolor{gray!60}32k & 36.06 & 18.16 & 14.96 & 11.79 & 24.92 & 12.09 & 19.66 \\

Longchat-7b-16k & \xmark &\cellcolor{gray!40}16k & 38.37 & 26.78 & 8.31 & \bf 15.17 & 20.21 & 9.74 & 19.76 \\
Vicuna1.5-7b-16k & \xmark &\cellcolor{gray!40}16k & 39.31 & 18.04 & 18.44 & 8.19 & 19.39 & 21.80 &  20.86 \\
Longchat-13b-16k & \xmark & \cellcolor{gray!40}16k & 37.85 & 21.11 & 12.18 & 14.76 & 22.75 & 14.95 & 20.60 \\
Vicuna1.5-13b-16k & \xmark &\cellcolor{gray!40}16k & \bf45.57 & 18.16 & 15.88 & 15.03 & 37.13 & 23.40 &  25.86 \\

Llama2-13b-NTK\color{blue}{*} & \xmark &\cellcolor{gray!40}16k & 30.99 & 15.88 & 13.61 & 6.89 & 11.13 & 15.58 &  15.67 \\
Llama2-13b-NTK(Dyn)\color{blue}{*} & \xmark &\cellcolor{gray!40}16k & 39.99 & 18.59 & 25.49 & 13.09 & 14.51 & \bf 26.90 & 23.09 \\

% longchat-7b-16k & \xmark & \cellcolor{gray!25}8k & 31.87 & 18.35 & 7.14 & 5.20 & 8.64 & 5.98  &  12.86\tiny{$\downarrow$} \\
Longchat-13b-16k & \xmark & \cellcolor{gray!25}8k & 36.94 & 16.70 & 10.77 & 7.55 & 14.14 & 9.91 & 16.00 \\
Chatglm2-6b-8k & \xmark & \cellcolor{gray!25}8k & 33.17 & 15.76 & 13.76 & 7.02 & 3.50 & 6.36 & 13.26 \\
XGen-7b-8k & \xmark & \cellcolor{gray!25}8k & 36.40 & 22.01 & 17.08 & 9.41 & 13.88 & 20.23  &  19.83 \\
MPT-7b-65k & \xmark & \cellcolor{gray!25}8k & 10.01 & 6.24 & 3.95 & 1.77 & 0.77 & 1.68  & 4.06 \\

\bottomrule
\end{tabular}
% \captionsetup{justification=centering}
\caption{Performance of various models on open-ended QA datasets in terms of F1 score. For results tested with N-gram metrics, please note that the results may not be accurate when the performance of the models is very similar or there is a large difference in the granularity of the output.}
\label{table:f1}
\end{table}

\begin{table*}[h!]
\centering \footnotesize
\renewcommand\arraystretch{1.1}
\tabcolsep0.05 in
\begin{tabular}{lcccccccccccccc}
\toprule
\multicolumn{1}{c}{\multirow{2}[1]{*}{\textbf{Model}}} &
\multicolumn{1}{c}{\multirow{2}[1]{*}{\textbf{Tokens}}} &
\multicolumn{3}{c}{\textbf{paper\_assistant}} &
\multicolumn{3}{c}{\textbf{review\_summ}} &
\multicolumn{3}{c}{\textbf{meeting\_summ}} &
\multicolumn{1}{c}{\multirow{2}[1]{*}{\textbf{Avg}}} \\
 & & \textbf{R-1} & \textbf{R-2} & \textbf{R-L} &
 \textbf{R-1} & \textbf{R-2} & \textbf{R-L} &
 \textbf{R-1} & \textbf{R-2} & \textbf{R-L} & \\
\cmidrule(lr){1-1} \cmidrule(lr){2-2} \cmidrule(lr){3-5} \cmidrule(lr){6-8} \cmidrule(lr){9-11} 
Turbo-16k-0613 & \cellcolor{gray!40}16k & 39.55 & 10.92 & 18.61 & \textbf{30.18} & \textbf{7.14} & 18.67 & 30.20 & 7.22 & 19.31 & 20.20 \\
AdaEmb-Turbo-0613  &  \cellcolor{gray!15}4k & 38.07 & 9.61 & 17.33 & 29.81 & 6.47 & \textbf{18.91} & \textbf{31.92} & 8.24 & \textbf{20.84} &20.13\\
BM25-Turbo-0613 & \cellcolor{gray!15}4k & \textbf{41.59} & \textbf{13.39} & \textbf{21.24} & 29.89 & 5.99 & 18.19 & 31.37 & \textbf{8.50} & 20.65 & \textbf{21.20} \\
\midrule
\rowcolor{mypink!30}
\multicolumn{12}{c}{\textit{Truncating input tokens to the pretraining context length}} \\
\midrule
Llama2-7b-chat & \cellcolor{gray!15}4k &37.15& 9.47 & 18.05 &{29.75}& {6.61} & {18.96} & {28.75} & {6.24} & {19.37} & 19.37\\
Llama2-13b-chat & \cellcolor{gray!15}4k & 37.27 & 9.79 & 18.49 & \textbf{30.49} & \textbf{6.69} & \textbf{19.23} &\textbf{29.63} &\textbf{6.54} & \textbf{19.65} & \textbf{19.75} \\
Vicuna1.3-7b & \cellcolor{gray!5}2k & 34.63 & 8.73 & 16.87 &29.01 & 6.28 & 18.18 & 24.18 & 4.93 & 15.93 & 17.63\\
Longchat-7b-16k & \cellcolor{gray!5}2k  & {37.01} & {9.61} & {18.21} & 26.45 & 5.05 & 16.88 & 23.92& 4.65 & 15.75 & 17.50\\
Chatglm2-6b-8k & \cellcolor{gray!5}2k & {36.91} & {9.45} & {17.96} &27.74& 5.77& 17.62 & 25.92& 5.61&17.57 & 18.28\\
XGen-7b (2k-4k-8k)  & \cellcolor{gray!5}2k & \textbf{37.72} & \textbf{9.97} & \textbf{18.77} & {28.21} & {5.94} & {18.69} & {26.94} & {5.92} & {18.24} & 18.93\\

\midrule
\rowcolor{mypink!30}
\multicolumn{12}{c}{\textit{Truncating input tokens to the further finetuning context length}} \\
\midrule
Chatglm-6b-32k &  \cellcolor{gray!60}32k & 32.65 & 8.09 & 16.51& 22.05 & 6.10 & 16.61& \bf 28.94 &\bf 8.86 &\bf 20.83 &  17.84 \\
Longchat1.5-7b-32k &  \cellcolor{gray!60}32k & 32.49 & 7.79 & 15.97 &27.53 & 5.80 & 17.94& 25.29 & 5.22 & 16.49 & 17.16 \\

Longchat-7b-16k &  \cellcolor{gray!40}16k & 35.05 & 8.57 & 16.70&26.07& 5.97 & 17.06& 20.13& 4.74 & 13.21 & 16.38 \\
Vicuna1.5-7b-16k &  \cellcolor{gray!40}16k & 36.84 & 9.78 & 17.66&28.91 & 6.47& 18.25& 26.90 & 5.53 &17.33& 18.63 \\

Longchat-13b-16k & \cellcolor{gray!40}16k & 34.41 & 8.07 & 16.45 &27.24 & 5.63 & 17.00 & 24.58 &  5.85 & 16.32 & 17.28\\
Vicuna1.5-13b-16k & \cellcolor{gray!40}16k & 36.30 & 8.69& 18.20 &28.59 & 6.15 & 18.49 & 27.82& 6.39 & 18.83& 18.82\\
Llama2-13b-NTK\color{blue}{*} & \cellcolor{gray!40}16k & 35.22 & 8.53 & 17.04& 23.97 & 4.72 & 14.89& 18.92 & 4.13 & 13.16& 15.61 \\
Llama2-13b-NTK(Dyn)\color{blue}{*} & \cellcolor{gray!40}16k & 28.89 & 7.21 & 14.83 &26.86 & 5.33& 17.55 & 22.29 & 4.88 & 15.29 & 15.90\\

Longchat-13b-16k & \cellcolor{gray!25}8k & 34.29 & 8.21 & 16.06 &26.76 & 5.61 & 16.77 & 20.86 & 4.01& 13.81 & 16.26\\
% longchat-7b-16k & \cellcolor{gray!25}8k & 33.93 &8.22 & 16.32 & 23.19 & 5.19 & 15.09& 15.65 & 2.89 & 10.73 & 14.57\tiny{$\downarrow$} \\
Chatglm2-6b-8k & \cellcolor{gray!25}8k & \bf38.07 & \bf9.61 & 17.33 & \bf29.81 & \bf6.47 & 18.91 & 24.74 & 4.45 & 4.44 & 18.36\\
XGen-7b-8k  & \cellcolor{gray!25}8k & 35.94& 8.49 & \bf17.92 & 28.92 & 6.28 & \bf{19.11} & 28.06 & 6.12 & 19.17 & \bf 18.89\\
MPT-7b-65k & \cellcolor{gray!25}8k & 15.91 & 2.91 & 11.18& 7.66 & 1.00 & 7.00 & 5.24 & 0.71 & 5.10 & 6.30 \\
\bottomrule
\end{tabular}
\caption{Performance of various models on \textbf{query-based} summarization and generation tasks in terms of ROUGE. }
\label{table:rouge2}
\end{table*}

\begin{table*}[h!]
\centering \footnotesize
\renewcommand\arraystretch{1.1}
\tabcolsep0.04 in
\begin{tabular}{lcccccccccccccc}
\toprule
\multicolumn{1}{c}{\multirow{2}[1]{*}{\textbf{Model}}} &
\multicolumn{1}{c}{\multirow{2}[1]{*}{\textbf{Tokens}}} &
\multicolumn{3}{c}{\textbf{gov\_report}} &
\multicolumn{3}{c}{\textbf{news}} &
\multicolumn{3}{c}{\textbf{patent}} &
\multicolumn{3}{c}{\textbf{tv\_show}} &
\multicolumn{1}{c}{\multirow{2}[1]{*}{\textbf{Avg}}} \\
 & & \textbf{R-1} & \textbf{R-2} & \textbf{R-L} &
 \textbf{R-1} & \textbf{R-2} & \textbf{R-L} &
 \textbf{R-1} & \textbf{R-2} & \textbf{R-L} &
 \textbf{R-1} & \textbf{R-2} & \textbf{R-L} & \\
\cmidrule(lr){1-1} \cmidrule(lr){2-2} \cmidrule(lr){3-5} \cmidrule(lr){6-8} \cmidrule(lr){9-11} 
\cmidrule(lr){12-14} 
Turbo-16k-0613 & \cellcolor{gray!40}16k & \textbf{45.9} & \textbf{15.6} & \textbf{23.6} & 35.3 & 8.1 & 16.1 & \textbf{46.0} & \textbf{20.3} & \textbf{29.3} & \textbf{32.0} & \textbf{5.4} & \textbf{16.9} & \textbf{24.5}\\
AdaEmb-Turbo-0613 & \cellcolor{gray!15}4k & 45.0 & 14.3 & 20.8 & 35.7 & 7.7 & 15.4 & 45.6 & 15.9 & 27.6 & 30.0 & 3.3 & 15.2 & 23.0\\
bm25-Turbo-0613  & \cellcolor{gray!15}4k & 44.6 & 14.2 & 21.5 & \textbf{38.4} & \textbf{9.1} & \textbf{16.8} & 43.3 & 15.5 & 27.1 & 31.0 & 4.6 & 15.4 & 23.4\\

\midrule
\rowcolor{mypink!30}
\multicolumn{15}{c}{\textit{Truncating input tokens to the pretraining context length}} \\
\midrule

llama2-7b-chat & \cellcolor{gray!15}4k & 43.7 & 15.3 & 22.2& 33.2& 6.4 & 15.5 &  {49.2} & {22.9} & {31.6} & 29.4 & {4.8} & {15.6}& {24.1} \\
llama2-13b-chat & \cellcolor{gray!15}4k & \bf46.3 & 16.1 & 24.0 & \textbf{34.9} & \textbf{8.1} & \textbf{16.3} &48.4 & 20.9 & 30.3 & \textbf{32.6}  & \textbf{6.6} & \textbf{17.2} & \textbf{25.1} \\
vicuna1.3-7b & \cellcolor{gray!5}2k & 44.6 & 16.4 & 23.2 & 32.9& 6.9 &14.8 & 44.7 & 20.4 & 28.8& 28.7 & 3.6 & 14.8& 23.3 \\
longchat-7b-16k & \cellcolor{gray!5}2k &43.6 &16.2 &  23.7 & 28.1 & 4.8 & 13.0 & {47.0} & {22.2} & {30.9} & 27.2 & 3.0 &14.4 & 22.8 \\
chatglm2-6b-8k & \cellcolor{gray!5}2k & {45.2} & \textbf{18.3} & \textbf{24.6} & {32.1} & {6.9} & {15.0} & 44.6 & 22.1 & 30.0 & 26.4& 2.6 &13.8  & 23.4\\
xgen-7b-8k & \cellcolor{gray!5}2k & 45.1 & 17.2 & 22.9 & 35.0 & 7.5 & 15.5 & \textbf{49.6} & \textbf{25.2} & \textbf{34.6}& {28.8} & {3.6} & {15.4} & 23.9\\

\midrule
\rowcolor{mypink!30}
\multicolumn{15}{c}{\textit{Truncating input tokens to the further finetuning context length}} \\
\midrule
chatglm2-6b-32k & \cellcolor{gray!60}32k & 38.1& 16.1 & 21.0 &24.2 &5.8 & 12.8 & 46.5 & 24.1 & \bf32.5& 23.4 & 4.2 & 13.8 & 21.8\\
longchat1.5-7b-32k & \cellcolor{gray!60}32k & 45.7 & 17.7 & 24.0 &\bf 36.8 &\bf 8.7 & 15.7 &42.0 & 18.2 & 27.2& 21.5 & 2.7 & 13.0 & 22.7\\

longchat-7b-16k & \cellcolor{gray!40}16k & 47.2 & 18.9 &23.9 & 27.7 & 5.4 & 13.4 & 46.2 & 20.9 & 30.1& 26.2 & 3.3 & 14.7 & 23.1\\
vicuna1.5-7b-16k & \cellcolor{gray!40}16k &47.2 & 18.9 & 25.0 & 32.3 & 6.8 & 15.5 & \bf 48.1 &\bf 25.1 & 32.4 & 26.0 & 3.6 & 14.8 & \bf 24.6  \\

longchat-13b-16k & \cellcolor{gray!40}16k & 46.2 & 18.2 & 24.1 & {35.2} & {7.6} & 15.8 &45.3 & 22.6 & 29.8& \bf31.9 & \bf6.0 &\bf17.3 & 24.0 \\
vicuna1.5-13b-16k & \cellcolor{gray!40}16k & 45.2 & 17.9 & 24.2& 31.6 & 6.8 & 15.2 &46.1 & 21.8 & 30.0 & 28.3 & 3.7 & 16.3 & 23.9  \\
llama2-13b-NTK & \cellcolor{gray!40}16k & 33.0 & 11.0 & 17.7 & 26.0 & 6.4 &  13.5 &37.9 & 13.5 & 22.9& 25.6 & 5.3 & 14.0 & 18.9  \\
llama2-13b-NTK(Dyn) & \cellcolor{gray!40}16k & 42.0 & 14.9 & 22.4& 34.0 & 7.8 & \bf15.9 & 45.3 & 19.1 & 28.5 & 25.5 & 3.9 & 13.9 & 22.7  \\

% longchat-7b-16k & \cellcolor{gray!25}8k &46.2 & 18.7 & 24.0 &30.4 & 6.9 & 13.5 & 41.9 & 20.0 & 27.1 & 26.2 & 4.7& 14.2  & 22.8\tiny{$\downarrow$}\\
longchat-13b-16k & \cellcolor{gray!25}8k & \textbf{49.3}&\textbf{19.5} &\textbf{25.1}& {34.9}& {7.4} & 15.5 & 43.5 & 20.1 & 28.0 & {31.0} & 4.5 & 15.7 & 24.5\\
chatglm2-6b & \cellcolor{gray!25}8k & 40.6 & 14.3 & 21.5 & 32.9 & 7.2 & 15.1 & 46.3 & 22.3 & 31.4 & 27.5 & 2.6 &14.5 &23.0\\
xgen-7b-8k & \cellcolor{gray!25}8k &40.2 & 13.8 & 21.1 & 31.9 & 6.0 & 15.3& 45.9 & 21.4 & 29.2 & 28.2 & 3.3 & 15.2 & 22.6 \\
mpt-7b-65k & \cellcolor{gray!25}8k &33.3 &10.7 & 19.3 &13.6&1.5 &9.2 & 25.5 & 12.2& 20.2 & 11.0 & 1.3 &6.4 & 13.6\\

\bottomrule
\end{tabular}
\caption{Performance of various models on long document summarization tasks in terms of ROUGE.}
\label{table:rouge1}
\end{table*}

\paragraph{Results from n-gram metrics}
Test all cases in open-ended tasks in L-Eval with GPT-4 is affordable. To give an overview of all the models on open-ended tasks, we test all models with n-gram metrics. As can be seen from the win rate from LLM judges (Table~\ref{tab:llm_eval}) and human evaluation (Table~\ref{tab:human_eval}), there is still a significant margin between commercial LLMs and open-source LLMs. However, the margin is not clear enough based on n-gram metrics.
Based on n-gram metrics, the open-source LCLMs also fail to beat their origin short-context model on truncated context. Overall, current open-source LCLMs generally excel more in conventional \textbf{summarization tasks} that involve instructions like ``\textit{Summarize this document}'' compared with query-based summarization and QA tasks. As for query-based tasks that pose questions from a specific perspective, performance can be significantly degraded if the instruction isn't fully understood. As we mentioned before, the increased input length can also lower the model's ability to comprehend lengthy instructions, thereby inhibiting its capability to generate answers that closely match the length of the ground truth. This phenomenon is less likely to be observed with more sophisticated LLMs (i.e. Turbo-16k). A naive solution is adding the instruction at both the beginning and end of the long input but there is still room to improve the ability of instruction understanding for LCLMs.

\paragraph{Retrieve-based models vs long context models}  We compare a representative LCLM baseline Turbo-16k-0613  with its short version Turbo-4k-0613 but enhanced with retrieval in Table~\ref{table:acc_exam}(closed-ended tasks) and Table~\ref{tab:llm_eval}(open-ended tasks). We use a sparse retrieval retriever bm25 and a strong dense retriever text-embedding-ada-002. Retrieve-based approaches generally yield better outcomes for tasks that have readily retrievable answers. For example, for long lectures understanding where the long document always contains some definitions and explanations for some academic 
terms, retrieval-based approaches obtain better results. However, retrieval is not a general solution as its performance is strongly related to instruction and document style. For example, they would never answer questions like \textit{how many sentences are there in a document.}  
Our results show that \textbf{CodeU} and \textbf{GSM(16-shot)} in L-Eval can not be solved by retrieval.  Retrieval-based methods also face difficulties in automatically \textbf{identifying the query} from user inputs.
Retrieval methods demonstrate comparatively less satisfactory performance in tasks where the answer cannot be retrieved, such as topic retrieval or tasks that demand models with long-range reasoning abilities like financial QA. Retrieve-based models produce similar or even superior results for summarization tasks. This may be because some paragraphs resembling summaries can be retrieved.   Besides, we also noticed that the main reason why regular Turbo-0613 outperforms Turbo-16k is its superior ability to accurately follow instructions.  However, even for these tasks, there are instances where the predicted answer might be ``I don't know" or ``not mentioned" due to the limitation of the retrieval process.
When evaluating retrievers, bm25 often matches the performance of the dense retriever, ada-embedding, in closed-ended tasks. However, in the open-ended tasks, the dense retriever ada-embedding outperforms BM25 by more than two points. This superior performance can be attributed to the dense retriever's ability to leverage not only term matching but also semantic matching.

\paragraph{Dynamic NTK scaling rules do not hold in practical tasks}  
\begin{wrapfigure}{r}{0.4\textwidth}
  \centering
  \vspace{-3mm}
  \includegraphics[width=0.4\textwidth]{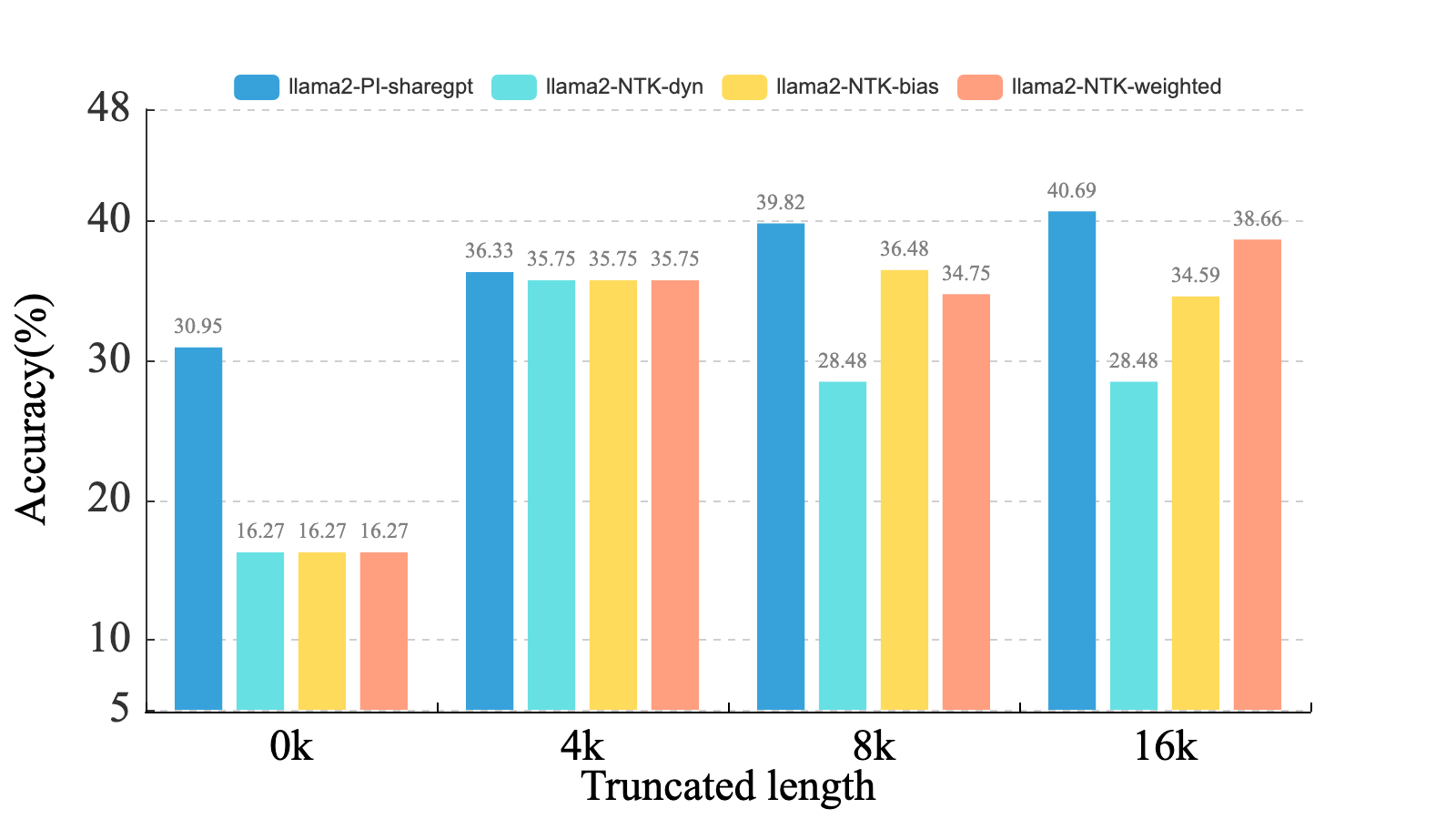}
  \caption{Peformance of different NTK-based methods when tackling input length at multiple scales.  }
  \label{fig:ntk-change}
  \vspace{-2mm}
\end{wrapfigure}
Dynamic NTK-aware positional embedding~\cite{dynamicNTK} is becoming more and more popular for extrapolation without further training. Based on dynamic NTK, given an input sequence with length $L$ and the model pertaining length $l$, we can set the original base 10,000 in RoPE to $10,000 \times \frac{L}{l}^{\frac{d}{d-2}}$ where $d$ is the head dimension, if we want to adapt the model to the longer context length $L$. We find that the scaling rule does not hold in practical tasks when the number of input tokens changes. The improvements can be further improved if using some variants of NTK. We study 2 simple modifications on the original dynamic NTK: (1) NTK-bias which means we use the base $10,000 \times (\frac{L}{l} + 1) ^{\frac{d}{d-2}}$ where 1 is the bias  (2) NTK-weighted which means we use the base $10,000 \times (\frac{L}{l} * 2) ^{\frac{d}{d-2}}$. Results are shown in Figure~\ref{fig:ntk-change} where Llama2-PI-sharegpt is a fine-tuned baseline using position interpolation. We test the results of 4 models by truncating the input length of test cases in Coursera. We can observe that employing which variants of NTK are strongly affected by the maximum tokens of the dataset.  When the input length is between 4k and 8k, NTK+bias gets the best results and NTK-weighted baseline is more robust on 16k input tokens. 

\begin{figure}
    \centering
    \includegraphics[width=1.0\textwidth]{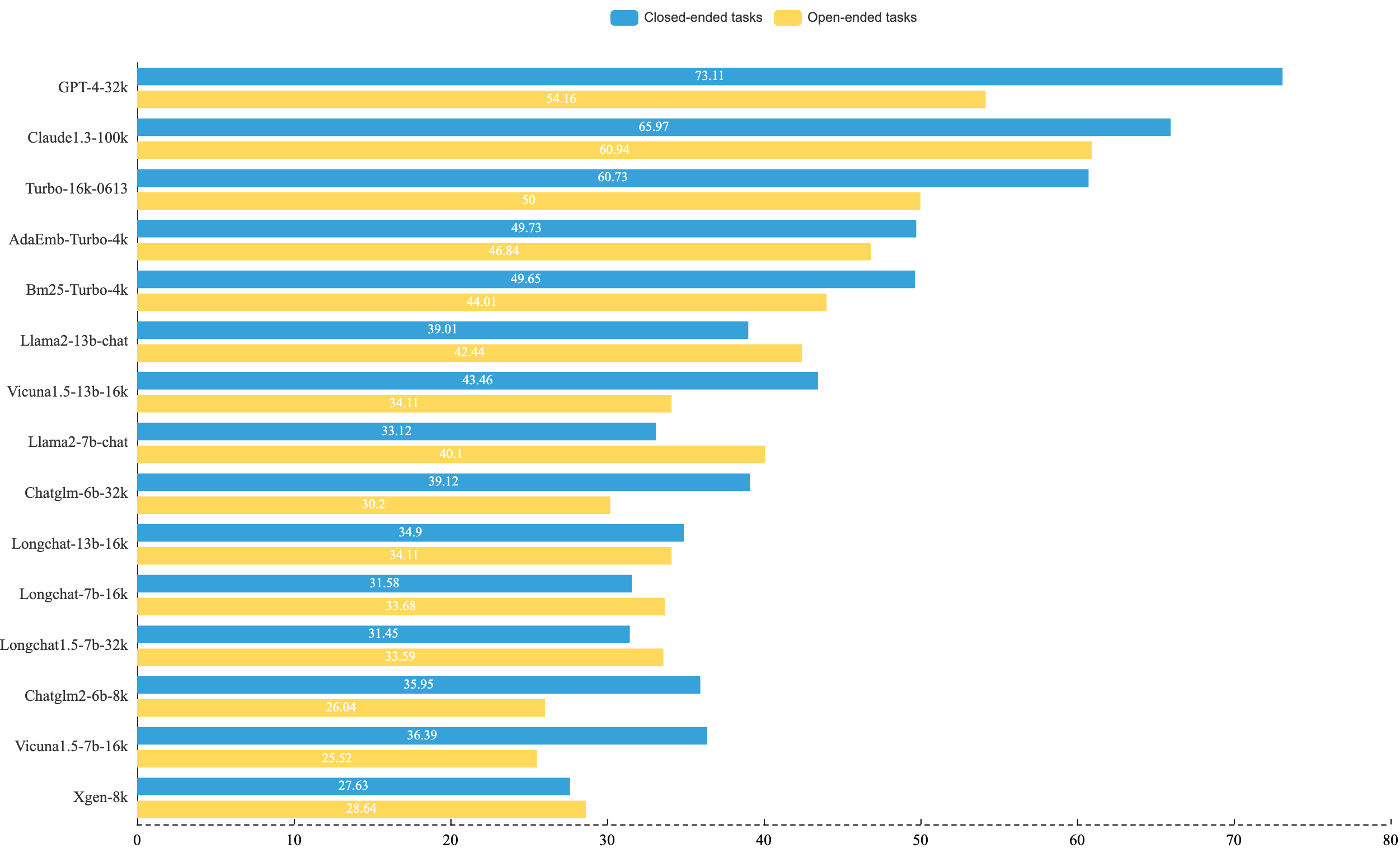}
    \caption{Overall results on open-ended tasks and closed-ended tasks. We find that GPT-4-32k is more capable of closed-ended tasks demonstrating powerful reasoning ability over long context since most closed-ended task in L-Eval has less than 32k input tokens, but the 100k context length help Cluade surpass both GPT-4-32k and Turbo-16k on open-ended tasks which generally has more input tokens. }
    \label{fig:overall}
\end{figure}

\begin{figure}
    \centering
    \includegraphics[width=1.0\textwidth]{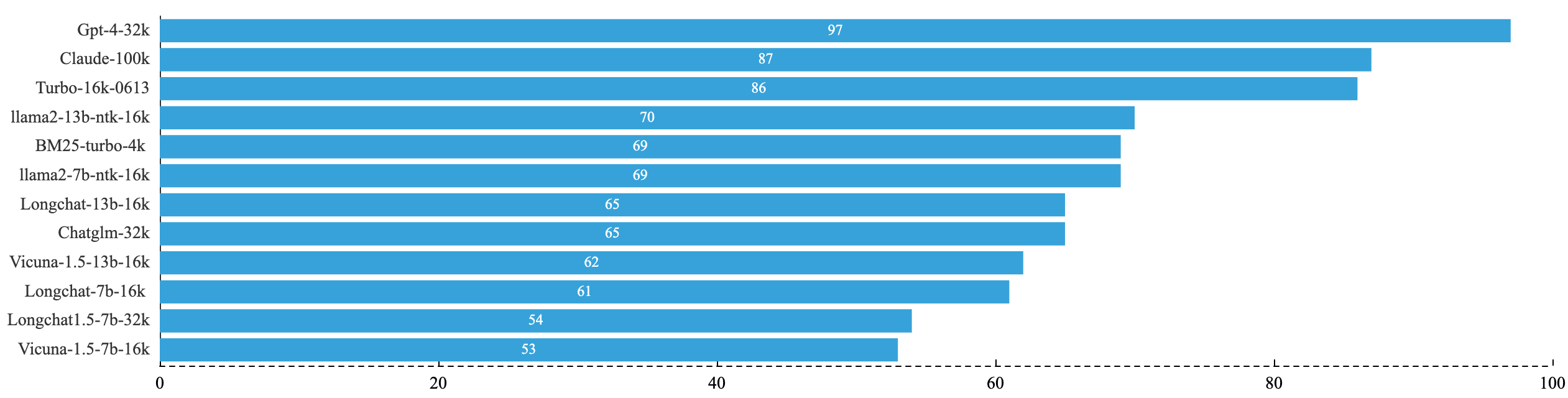}
    \caption{Overall results on the topic retrieval tasks. Testing short context models on this task with truncated input texts is unfair, so we only include long context LLMs.}
    \label{fig:topic_all}
\end{figure}

\newpage
\section{Data Collection and Annotation for L-Eval}
\label{sec:data-appendix}
In our pursuit of diverse, comprehensive, and relevant data, we sourced datasets from a wide array of platforms and sources. These datasets, represent various facets of everyday life and specialized fields and present different challenges for LCLMs. We leveraged resources from previous open-source datasets, Coursera subtitles, earning call transcripts from corporate websites, GitHub, etc. The instruction styles in L-Eval include multiple-choice questions, school math with many examples, key topics retrieval from lengthy dialogues, text summarization, and abstractive question answering, encompassing a wide range of tasks. The construction of each dataset and our effort to make it more challenging are as follows.

\subsection{TOFEL (English tests)} 
This dataset is sourced from the TOEFL Practice Online and we collect the data from TOEFL-QA~\citep{tseng2016towards,chung2018supervised} and all lectures from a single TPO have been consolidated into one lengthy lecture. After the consolidation, we select the top 15 longest lectures.
\begin{example}{}{example}
\texttt{\scriptsize
\textbf{Input}: <Multiple long lectures> \textbackslash n\textbackslash n\\
Question: why did Frantzen go to the sales barn\\A. to study human form and movement\\B. to earn money by painting portraits\\C. to paint farm animals in an outdoor setting\\D. to meet people who could model for her painting \\
\textbackslash n\textbackslash n Answer: \\
\textbf{Ground truth}: A
}
\end{example}

\subsection{GSM(16-shot)(Grade school math)}\label{sec:app-gsm}
This dataset is derived from 100-grade school math problems in the GSM8k dataset~\citep{cobbe2021training}. Increasing the number of high-quality and complex examples usually has a positive effect on solving math problems. We construct 16 in-context examples with lengthy Chain-of-thought for this task where 8 examples come from  \textit{chain-of-thought-hub}\footnote{\url{https://github.com/FranxYao/chain-of-thought-hub/blob/main/gsm8k/lib_prompt/prompt_hardest.txt}} using the hardest prompt and the remaining 8 examples are constructed by us. We selected 8 questions from GSM8k based on their difficulty and annotated the solving process.  Models with 2k or 4k context length face difficulties while encoding the 16 examples. We experiment with the newly constructed examples and it perform better than only encoding 8 examples. Concretely, the accuracy rises from 79 (8-shot) to 84 (16-shot) when using turbo-16k-0613 as the base model. 
\begin{example}{}{example}
\texttt{\scriptsize
\textbf{Input}: <example 1>  \textbackslash n\textbackslash n <example 2> \textbackslash n\textbackslash n  ... <example n> \textbackslash n\textbackslash n \\
Question: Janet’s ducks lay 16 eggs per day. She eats three for breakfast every morning and bakes muffins for her friends every day with four. She sells the remainder at the farmers' market daily for \$2 per fresh duck egg. How much in dollars does she make every day at the farmers' market? \textbackslash n\textbackslash n \\
Let's think step by step\\
\textbf{Ground truth}: 18
}
\end{example}

\subsection{QuALITY (Gutenberg)}\label{sec:app-quality}
This dataset 
is sourced from the multiple choice QA dataset QuALITY~\citep{pang2022quality} which contains multiple-choice questions derived from the literature on Gutenberg. We filter 20 long stories and 202 questions and correct/delete questions with annotation errors. We found that most questions in QuALITY can be solved by extracting paragraphs from long texts. We further enhance some synthesis questions that need a global understanding of the document. Examples of the annotated synthesis questions are as follows:
\begin{enumerate}[itemsep=0.1ex,leftmargin=2em]
    \item \textit{What can we infer from the longest sentence in the story?}
    \item \textit{The longest dialogue is spoken by whom?}
    \item \textit{Extract names mentioned in the longest sentence in the story.}
    \item \textit{How many words are there in the story?}
    \item \textit{How many sentences are there in the story?}

\end{enumerate}
The reference source sentences are automatically located and the ground truth answers are manually annotated by us. An example of the original question in QuALITY is like this:

\begin{example}{}{example}
\texttt{\scriptsize
\textbf{Input}: <A long story>\textbackslash n\textbackslash n \\
{Instruction}: Why did Syme accept the mission with Tate?\\ (A) He needed a way back to Earth\\ (B) He felt he would collect a reward along the way\\ (C) He respected Tate\\ (D) He had no plan for his life, so he jumped on the adventure\\
\textbf{Ground truth}: (B) He felt he would collect a reward along the way
}
\end{example}

\subsection{Coursera (Advanced lectures)}\label{sec:coursera}
This dataset originates from the Coursera website\footnote{\url{https://coursera.org/}}. We selected and completed 4 courses: 
\begin{enumerate}[itemsep=0.1ex,leftmargin=2em]
    \item \textit{Ask Questions to Make Data-Driven Decisions},
    \item \textit{Data Scientist's Toolbox}, 
    \item \textit{Process data from dirty to clean}, 
    \item \textit{Improving Deep Neural Networks: Hyperparameter Tuning, Regularization and Optimization}.
\end{enumerate}
The input long document is the subtitles of the videos  and we merge courses in one week into one single long lecture. Questions and the ground truth answers are labeled by the authors. The instruction style of Coursera takes the format of multiple choice. In order to increase the difficulty of the task, we have set \textbf{multiple correct options}. Failure to select all correct choices will result in receiving only a quarter of the total points for that question.
\begin{example}{}{example}
\texttt{\scriptsize
    \textbf{Input}: <A long lecture>\textbackslash n\textbackslash n\\
    Question:  When working with a new team, which of the following actions can help you to adapt to different communication expectations? Select all that apply.\\
    A. Ask questions when you are unsure of something\\
    B. Learn the team's preferred communication style\\
    C. Observe how teammates communicate with each other\\
    D. Ignore the team's communication preferences and use your own style\\
    \textbackslash n\textbackslash n Answer:\\
    \textbf{Ground truth}: ABC 
}
\end{example}

\subsection{SFcition (Scientific fictions)}\label{sec:sfiction}
We annotate this sub-task to test the loyalty of the LCLM to the input context. LLMs have acquired many commonsense in their pertaining corpus known as parametric knowledge~\citep{wang2023easyedit}. However, we argue that in LCLMs, contextual knowledge is more crucial than parametric knowledge. In real-world applications, many long documents are private and can never be seen during pretraining. It may contain new knowledge or describe a new world which may be opposite to the pretraining knowledge. The language model should follow contextual knowledge instead of parametric knowledge. To simulate this scenario, we annotate a science fiction dataset consisting of True or False questions. The original works are sourced from SFGram\footnote{\url{https://github.com/nschaetti/SFGram-dataset}}. We manually select documents that fit our experimental conditions and annotate them with questions and corresponding answers. Most of the answers to these questions contradict real-world principles and do not comply with actual physical laws, such as the statement: \textit{Humans have invented the time machine}. As a result, open-source models have very serious hallucination problems which in turn help them acquire a high score on this dataset. So we also give the answer based on real-world knowledge, and the final accuracy is calculated by the average of loyalty and factuality.
\begin{example}{}{example}
\texttt{\scriptsize
\textbf{Input}: <A scientific fiction>\textbackslash n\textbackslash n \\
{Question}: We cannot get to the centre of the Earth, True or False? Answer this question based on the world described in the document. \\
\textbf{Ground truth}: False \\\\
Question: We cannot get to the centre of the Earth, True or False? Answer this question based on the real-world knowledge and facts up until your last training. \\
\textbf{Ground truth}: True
}
\end{example}

\subsection{CodeU (Python)}\label{sec:codeU}
This dataset is used to test the capability of understanding long code.  Given a lengthy code base, we will call some functions defined in the codebase and the model should infer the final output of the program. We mainly use source code from Numpy\footnote{\url{https://github.com/numpy/numpy}}. We also write a string processing codebase containing more than 100 functions that take a string as input such as extracting the email address from an input string. To prevent LLMs from answering the question based on their parametric knowledge, We replace the original function name defined in Numpy with \texttt{Op1, Op2..., OpN}. The Language Model (LLM) should first identify where the function is called and determine which functions are invoked, ultimately ascertaining the results of the operations. CodeU represents the most challenging task within L-Eval. Even the most potent model, GPT-4-32k, achieves an accuracy of only 25.55\%.
\begin{example}{}{example}
\texttt{\scriptsize
    \textbf{Input}: <The beginning of a lengthy Python program>  \\
    def Op1(): ...\\
    def Op2(): ... \\
    args = [4,5,6] \\
    output = Op1(args) \\
    print(output) \\
    <The rest of the program>\textbackslash n\textbackslash n \\
    Instruction: What is the output of this program? Please carefully read through these code snippets and comments. You should first identify where the functions are defined and then figure out what they do.\\
    \textbackslash n\textbackslash n let's think step by step:\\
    \textbf{Ground truth}: [1,2,3] 
}
\end{example}

\subsection{TopicRet (Lengthy conversation)} 
This dataset comes from the LongChat repository~\citep{longchat2023}\footnote{\url{https://github.com/DachengLi1/LongChat}}, and its task style focuses on retrieving topics from extensive chat histories. Recent studies show that language models are good at retrieving information from the very beginning or end of its input context but are usually lost in the middle~\citep{liu2023lost}. To make the task more challenging,  we enhance the original task by asking the model to extract \textbf{the second and the third} topic.

\begin{example}{}{example}
\texttt{\scriptsize
\textbf{Input}: <A long conversation > \textbackslash n\textbackslash n\\
Question: What is the second topic we discussed? Only give me the topic name. Do not summarize yourself.\\
\textbf{Ground truth}: The future of space tourism
}
\end{example}

% open-ended tasks

\subsection{LongFQA (Finance)}\label{sec:longfqa}
We find that there is a lack of long open-ended QA datasets in finance.  The long context finance dataset is derived from earnings call transcripts obtained from the \textit{Investor Relations} section of the company websites. We annoate 6 transcripts from 6 different incorporations, Lumentum Oclaro\footnote{\url{https://investor.lumentum.com/overview/default.aspx}}, Theragenics\footnote{\url{https://www.sec.gov/Archives/edgar/data/}}, FS KKR Capital Corp\footnote{\url{https://www.fskkradvisor.com/investor-relations/}}, LaSalle Incorporated\footnote{\url{https://ir.jll.com/overview/default.aspx}}, Renewable Energy Group\footnote{\url{https://www.regi.com/resources/press-releases}} with 54 questions based on these transcripts.
\begin{example}{}{example}
\texttt{\scriptsize
\textbf{Input}: <A long document>\textbackslash n\textbackslash n  \\
Instruction: You are asked to act as a member of the Financial Results Conference Call and answer the question: What major actions has Greg Dougherty, the CEO of Oclaro, highlighted as being undertaken by the company for its restructuring plan? \textbackslash n Answer this question with xx words.\\
\textbf{Ground truth}: Oclaro has been implementing a significant restructuring plan, which includes closing our second major...
}
\end{example}

\subsection{CUAD (Law)}
Questions on the Legal domain are drawn from the CUAD (Contract Understanding Atticus Dataset) dataset~\citep{hendrycks2021cuad} designed for supporting NLP research for automating legal contract review. We manually filter 20 documents with annotated QA pairs from CUAD.
\begin{example}{}{example}
\texttt{\scriptsize
\textbf{Input}: <Legal contracts> \textbackslash n\textbackslash n \\
Instruction: Highlight the parts (if any) of this contract related to \"Expiration Date\" that should be reviewed by a lawyer. Details: On what date will the contract's initial term expire? \textbackslash n Answer this question with xx words.\\
\textbf{Ground truth}: The term of this Agreement shall commence on the Effective Date and shall continue in full force and effect for an initial period of five (5) years.
}
\end{example}

\subsection{MultiDoc2Dial (Dialogues over multi-documents)} 
This dataset is sampled from the MultiDoc2Dial dataset~\citep{Feng_2021} which aims to model goal-oriented dialogues grounded in multiple documents. It contains dialogues from 4 different domains: Finance, Travel, Entertainment, and Shopping. Each dialogue in the dataset is grounded in 2-5 relevant documents covering different topics within the domain.
\begin{example}{}{example}
\texttt{\scriptsize
\textbf{Input}: <Multiple long documents> \textbackslash n\textbackslash n \\
Instruction: How long will Driver's Ed courses be valid for? \textbackslash n Answer this question with xx words.\\
\textbf{Ground truth}: For roughly 1 one year. Maybe longer depending on the course.
}
\end{example}

\subsection{Natural Questions (Wikipedia)} 
We filter 20  wikipedia long documents from  Natural Question~\citep{kwiatkowski-etal-2019-natural} on Google Research datasets.  Questions can be answered with the same documents are merged, and duplicate questions are removed.
\begin{example}{}{example}
\texttt{\scriptsize
\textbf{Input}: <Documents from Wiki>\textbackslash n\textbackslash n \\
Instruction: when did season 2 of handmaid's tale start? \textbackslash n Answer this question with xx words. \\
\textbf{Ground truth}: April 25, 2018
}
\end{example}

\subsection{NarrativeQA (Narratives)} This dataset is collected from NarrativeQA~\citep{kočiský2017narrativeqa} which has the longest document length in L-Eval. The original question-answering dataset was created using entire books from Project Gutenberg\footnote{\url{https://www.gutenberg.org}} and movie scripts from various websites. Summaries of the books and scripts were taken from Wikipedia and given to annotators. Our work focuses on correcting the annotation error for example, there are some issues where the main character in the question does not even appear in the input document at all. 
\begin{example}{}{example}
\texttt{\scriptsize
\textbf{Input}: <A long novel>\textbackslash n\textbackslash n  \\
Instruction: Why did Mary pay off the debt for Ann's family? \textbackslash n Answer this question with xx words.\\
\textbf{Ground truth}: Mary was in love with Ann.
}
\end{example}

\subsection{Qasper (Papers)} 
This dataset is filtered from the Qasper dataset~\citep{dasigi2021dataset}, which is a question-answering resource focused on NLP papers. The dataset was constructed using NLP papers that were extracted from the Semantic Scholar Open Research Corpus (S2ORC). After filtering, we remove the unanswerable questions and the extractive version answers. We also discovered instances where identical questions yielded contradictory answers. We addressed this issue by meticulously reviewing the paper and rectifying the incorrect responses.
\begin{example}{}{example}
\texttt{\scriptsize
\textbf{Input}: <A long paper>\textbackslash n\textbackslash n \\
Instruction: How did they obtain the dataset? \textbackslash n Answer this question with xx words.\\
\textbf{Ground truth}: public resources where suspicious Twitter accounts were annotated, list with another 32 Twitter accounts from BIBREF19 that are considered trustworthy.
}
\end{example}

\subsection{Openreview (Papers)} This task aims to help researchers working on scientific papers by dealing with tasks like correcting grammar errors or typos and writing some sections. We include 3 tasks in the paper writing assistant task of L-Eval: 1) writing an Abstract section, (2)  writing a Related Work section,  and (3) finally giving a review of this paper including valuable suggestions and questions. Notably, we discourage reviewers from using large models for reviews. Our aim is to assist authors in further improving their papers. Therefore, we ask the model to give some valuable suggestions and raise some questions for authors. We filter 20 papers with well-written reviews for L-Eval. We use the processed PDF files from~\citet{yuan2021automate}.
\begin{example}{}{example}
\texttt{\scriptsize
\textbf{Input}: <A long paper>\textbackslash n\textbackslash n \\
\textbf{1}. Instruction: Please generate the Abstract section for this paper. \textbackslash n Answer this question with xx words. \\
\textbf{2}. Instruction: Please summarize related work and you should include the following works [a list of papers]. \textbackslash n Answer this question with xx words.\\ 
\textbf{3}. Instruction: Please write a review for this paper and you should provide some suggestions and raise some questions in your review. \textbackslash n Answer this question with xx words.\\
\textbf{Ground truth}: Conventional out-of-distribution (OOD) detection schemes based on variational autoencoder or Random Network Distillation (RND) have been observed to assign ...
}
\end{example}

\subsection{GovReport (Government Reports)} 
This dataset is filtered  from the government report summarization dataset~\citep{huang-etal-2021-efficient}, the dataset consists of long reports written by U.S. government research agencies such as the Congressional Research Service and Government Accountability Office. The documents and summaries in this dataset are longer compared to other long document summarization datasets. We manually filter 13 documents with human-written summaries from the original dataset.
\begin{example}{}{example}
\texttt{\scriptsize
\textbf{Input}: <A government report>\textbackslash n\textbackslash n \\
Instruction: Please help me summarize this government report. \textbackslash n Answer this question with xx words.\\
\textbf{Ground truth}: The President of the United States has available certain powers that may be exercised in the event that the nation is threatened by crisis, exigency, or emergency circumstances...
}
\end{example}

\subsection{QMSum (Meetings)}
This dataset sourced from the QMSum~\citep{zhong2021qmsum}, this dataset contains query-based meeting summarizations.  Query-based summarization aims to summarize the document given a specific aspect. We selected 20 meeting transcripts accompanied by queries, specifically choosing those that could not be easily addressed through retrieval methods.
\begin{example}{}{example}
\texttt{\scriptsize
\textbf{Input}: <Meeting trancripts>\textbackslash n\textbackslash n \\
Instruction: What was agreed upon on sample transcripts? \textbackslash n Answer this question with xx words.\\
\textbf{Ground truth}: To save time, speaker mn005 will only mark the sample of transcribed data for regions of overlapping speech, as opposed to marking all acoustic events...
}
\end{example}

\subsection{SPACE (Reviews)} 
The review (opinion) summarization aims to summarize the reviews from customs reviews on a restaurant or hotel. We obtain 20 samples from the validation and test set of SPACE~\citep{angelidis-etal-2021-extractive} where human-written abstractive summaries are created for 50 hotels based on 100 input reviews each. SPACE consists of customer reviews of hotels from TripAdvisor, with 1.1 million training reviews for 11,000 hotels.  The original task asks the model to summarize hotels from multiple aspects: food, location, cleanliness, etc. We construct the instructions for review summarization with GPT-4 and some examples.
\begin{example}{}{example}
\texttt{\scriptsize
\textbf{Input}: <Multiple reviews>\textbackslash n\textbackslash n \\
Instruction: Give a broad summary of guest impressions about Doubletree by Hilton Seattle Airport. \textbackslash n Answer this question with xx words.\\
\textbf{Ground truth}: The staff are friendly and exceptional. Every room (lobby included) was very clean. They are spacious, very quiet, and come with a coffee maker...
}
\end{example}

\subsection{Multi-News (News)} This dataset is sourced from the Multi-News~\citep{fabbri2019multinews}. The original Multi-News dataset contains news articles as well as human-written summaries of those articles, compiled from the website newser.com  where each article consists of multiple short news articles. We select 10 articles for the L-Eval benchmark.
\begin{example}{}{example}
\texttt{\scriptsize
\textbf{Input}: <News articles>\textbackslash n\textbackslash n \\
Instruction: Please summarize these news articles. \textbackslash n Answer this question with xx words.\\
\textbf{Ground turth}:  Why did Microsoft buy Nokia's phone business? We now know Microsoft's answer: The computing giant released a 30-slide presentation today arguing that the move will improve Microsoft...
}
\end{example}

\subsection{BigPatent (Patents)} 
This dataset is derived from the BigPatent~\citep{sharma-etal-2019-bigpatent} project, which consists of 1.3 million records of U.S. patent documents along with human-written abstractive summaries, we select 13 patent patents from the original dataset. 
\begin{example}{}{example}
\texttt{\scriptsize
\textbf{Input}: <A long patent>\textbackslash n\textbackslash n \\
Instruction: You are a patent examiner. Please write a summary of this patent. \textbackslash n Answer this question with xx words.\\
\textbf{Ground truth}: The invention provides a method and system for cleaning pet paws by providing a bounded container containing...
}
\end{example}

\subsection{SummScreen (TV show)} 
This dataset originates from the SummScreen~\citep{chen2022summscreen}, the original dataset is an abstractive summarization dataset combining TV series transcripts and episode recaps. SummScreen is constructed from fan-contributed websites. We use 13  of these  transcripts in L-Eval.
\begin{example}{}{example}
\texttt{\scriptsize
\textbf{Input}: <TV series transcripts> \textbackslash n\textbackslash n \\
Instruction: Write a summary of the scene. \textbackslash n Answer this question with xx words.\\
\textbf{Ground turth}: Feeling guilty over Phoebe missing out on London, the gang plans a weekend trip to Atlantic City, but just as they are about to leave...
}
\end{example}

\end{document}